%% file: main.tex
\documentclass[nohyperref]{article}

\usepackage{microtype}
\usepackage{graphicx}
\usepackage{subcaption}
\usepackage{booktabs} 

\usepackage{hyperref}

\usepackage[accepted]{icml2022}

\usepackage[T1]{fontenc}
\usepackage[utf8]{inputenc}
\usepackage[scaled=.8]{beramono}
\usepackage[defaultsans,scale=.8]{opensans}

\usepackage{amsmath}
\usepackage{amssymb}
\usepackage{mathtools}
\usepackage{nicefrac}
\usepackage{enumitem}
\usepackage{bm}
\usepackage{framed}
\usepackage{xcolor}
\usepackage{xspace}
\usepackage{booktabs}
\usepackage{tikz}
\usepackage{array}
\usepackage{tikz-dependency}
\usepackage{multirow}
\usetikzlibrary{positioning,arrows}
\setlist[itemize,enumerate]{leftmargin=*}

\DeclareMathOperator{\conv}{conv}
\DeclareMathOperator*{\argmin}{argmin}
\DeclareMathOperator*{\argmax}{argmax}
\DeclareMathOperator*{\softmax}{softmax}

\DeclareMathOperator*{\relu}{relu}

\DeclareMathOperator*{\Tr}{Tr}

\newcommand{\entropy}{\mathcal{H}}

\newcommand\reals{\mathbb{R}}
\newcommand\DP[2]{\langle #1, #2 \rangle}
\newcommand\bigDP[2]{\left\langle #1, #2 \right\rangle}

\newcommand\Reg{\Psi}

\newcommand*{\eg}{\textit{e.\hspace{.07em}g.}\@\xspace}
\newcommand*{\ie}{\textit{i.\hspace{.07em}e.}\@\xspace}

\newcommand{\ltwo}[1]{\frac{1}{2}\|#1\|^2} 
\newcommand{\energyltwo}[1]{E_#1(#1) = \ltwo{#1}}

\newcommand\var[1]{\mathsf{#1}}
\newcommand\initit[1]{{{#1}^{(0)}}}
\newcommand\nextit[1]{{{#1}_\star}}
\newcommand\currit[1]{#1}

\usepackage{amsmath}
\usepackage{amssymb}
\usepackage{mathtools}
\usepackage{amsthm}

\usepackage[capitalize,nameinlink]{cleveref}

\theoremstyle{plain}
\newtheorem{theorem}{Theorem} 

\newtheorem{lemma}[theorem]{Lemma}

\theoremstyle{definition}

\theoremstyle{remark}

\usepackage[textsize=tiny]{todonotes}

\icmltitlerunning{Modeling Structure with Undirected Neural Networks}

\begin{document}

\twocolumn[
\icmltitle{Modeling Structure with Undirected Neural Networks}

\icmlsetsymbol{equal}{*}

\begin{icmlauthorlist}
\icmlauthor{Tsvetomila Mihaylova}{ist}
\icmlauthor{Vlad Niculae}{uva}
\icmlauthor{Andr\'{e} F.T.~Martins}{ist,lumlis,unbabel}
\end{icmlauthorlist}

\icmlaffiliation{ist}{Instituto de Telecomunica\c{c}\~{o}es, Instituto Superior T\'{e}cnico, Lisbon, Portugal}
\icmlaffiliation{uva}{Language Technology Lab, University of Amsterdam, The Netherlands}
\icmlaffiliation{lumlis}{LUMLIS, Lisbon ELLIS Unit, Portugal}
\icmlaffiliation{unbabel}{Unbabel, Lisbon, Portugal}

\icmlcorrespondingauthor{Tsvetomila Mihaylova}{tsvetomila.mihaylova@tecnico.ulisboa.pt}
\icmlkeywords{Machine Learning, ICML, Structure, Neural Networks, Energy Networks}

\vskip 0.3in
]



\printAffiliationsAndNotice{}  

\begin{abstract}
Neural networks are powerful function estimators, leading to their status as a paradigm of choice for modeling structured data.
%
However, unlike other structured representations that emphasize the modularity of the problem --
\textit{e.g.}, factor graphs -- neural networks are usually monolithic mappings
from inputs to outputs, with a fixed computation order. This limitation prevents them from
capturing different directions of computation and interaction between the modeled variables.
In this paper, we combine the representational strengths of factor graphs and of neural networks, proposing \textit{undirected neural networks (UNNs)}:
a flexible framework for specifying computations that  can be performed in any order.
For particular choices,
our proposed models subsume and extend many existing architectures: feed-forward, recurrent, self-attention networks, auto-encoders, and networks with implicit layers.
We demonstrate
the effectiveness of undirected neural architectures, both unstructured and structured, on a range of tasks:
tree-constrained dependency parsing, convolutional image classification, and sequence completion with attention.
By varying the computation order, we show how a single UNN can be used both as a classifier and a prototype generator, and how it can fill in missing parts of an input sequence,
making them a promising field for further research.
\end{abstract}

\section{Introduction}
\input{introduction}

\input{unn}

\input{experiments_mnist}
\input{experiments_attention}
\input{experiments_parsing}

\input{related_work}

\input{conclusions}
\input{acknowledgements}

\bibliography{custom}
\bibliographystyle{icml2022}


\include{appendices}


\end{document}

%% file: introduction.tex

\looseness=-1
Factor graphs have historically been a very appealing toolbox for representing structured prediction problems \citep{bakir2007predicting,smith2011linguistic,nowozin2014advanced}, with wide applications to vision and natural language processing applications.  
In recent years, neural networks have taken over as the model of choice for tackling many such applications. Unlike factor graphs\,--\,which emphasize the modularity of the problem\,--\,neural networks typically work end-to-end, relying on rich representations captured at the encoder level (often pre-trained), which are then propagated to a task-specific decoder. 

In this paper, we combine the representational strengths of factor graphs and neural networks by proposing \textbf{undirected neural networks} (UNNs) -- a framework in which outputs are not computed by evaluating a composition of functions in a given order, but are rather obtained \textit{implicitly} by minimizing an energy function which factors over a graph. 
For particular choices of factor potentials, UNNs subsume many existing architectures, including feedforward, recurrent, and self-attention neural networks, auto-encoders, and networks with implicit layers. 
When coupled with a coordinate descent algorithm to minimize the energy, the computation performed by an UNN is similar (but not equivalent) to a neural network sharing parameters across multiple identical layers. 
Since UNNs have no prescribed computation order, the exact same network can be used to predict any group of variables (outputs) given another group of variables (inputs), or vice-versa (\textit{i.e.}, inputs from outputs) enabling new kinds of joint models. 
%
%
%
%
%
In sum, our contributions are:

\begin{itemize}
    \item We present UNNs and show how they extend many existing neural architectures.   
    \item We provide a coordinate descent inference algorithm, which, by an
    ``unrolling lemma'' (\cref{lemma:unrolling}),
    can reuse current building blocks from feed-forward networks in a modular way.
    \item We develop and experiment with multiple factor graph architectures, tackling both structured and unstructured tasks, such as natural language parsing, image classification, and image prototype generation. We develop a new undirected attention mechanism and demonstrate its suitability for sequence completion.\footnote{The source code is on \href{https://github.com/deep-spin/unn}{https://github.com/deep-spin/unn}.}
\end{itemize}




\paragraph{Notation.}
We denote vector values as $a$, matrix and tensor values as $A$, and abstract factor graph variables as $\var{A}$.
The Frobenius inner product of two tensors with matching dimensions $A, B \in \reals^{d_1 \times \ldots \times d_n}$ is
$\DP{A}{B} \coloneqq \sum_{i_1=1}^{d_1} \ldots \sum_{i_n=1}^{d_n} a_{i_1\ldots{}i_n} b_{i_1 \ldots i_n}$. For vectors
$\DP{a}{b}=a^\top b$ and for matrices $\DP{A}{B}=\Tr(A^\top B)$.
The Frobenius norm is $\|A\| \coloneqq \sqrt{\DP{A}{A}}$.
Given two tensors $A \in \reals^{c_1, \ldots, c_m}, B \in \reals^{d_1, \ldots, d_n}$, their outer product is
$(A \otimes B)_{i_1, \ldots, i_m, j_1, \ldots, j_n} = a_{i_1, \ldots, i_m} b_{j_1, \ldots, j_m}$. For vectors, $a \otimes b = ab^\top$.
We denote the $d$-dimensional vector of ones as $1_d$ (or tensor, if $d$ is a tuple.)
The Fenchel conjugate of a function $\Psi : \reals^d \rightarrow \reals$ is
$\Psi^*(t) \coloneqq \sup_{x \in \reals^d} \DP{x}{t} - \Psi(x)$. When $\Psi$ is strictly convex,
$\Psi^*$ is differentiable, and
$(\nabla\Psi^*)(t)$ is the unique maximizer
$\argmax_{x \in \reals^d} \DP{x}{t} - \Psi(x)$. 
We denote the non-negative reals as
$\reals^d_+ \coloneqq \{ x \in \reals^d : x \geq 0 \}$,
and the $(d-1)$-dimensional simplex as
$\triangle_d \coloneqq \{ \alpha \in \reals^{d}_+, \DP{1_d}{\alpha}=1\}.$
The Shannon entropy of a discrete distribution $y \in \triangle_d$ is
$\entropy(y) \coloneqq -\sum_i y_i \log y_i$.
The indicator function 
$\iota_{\mathcal{X}}$ is defined as
$\iota_{\mathcal{X}}(x) \coloneqq 0$ if $x \in \mathcal{X}$, and $\iota_{\mathcal{X}}(x) \coloneqq +\infty$ otherwise.


%% file: unn.tex
\section{Undirected Neural Networks}\label{section:unn}



Let $\mathcal{G} = (V, F)$ be a factor graph, \textit{i.e.}, a bipartite graph consisting of a set of variable nodes $V$ and a set of factor nodes $F$, where each factor node $f \in F \subseteq 2^V$ is linked to a subset of variable nodes.
%
%
Each variable node $\var{X} \in V$ is associated with a representation vector $x \in \mathbb{R}^{d_\var{X}}$. 
We define unary energies for each variable
$E_\var{X}(x)$,
as well as higher-order energies 
$E_f(x_f)$, where $x_f$ denotes the values of all variables linked to factor
$f$. Then, an assignment defines a total energy function 
\begin{equation}\label{eq:energy}
E(x_1, \ldots, x_n) \coloneqq \sum_i
E_{\var{X}_i}(x_i) + \sum_f E_f(x_f).
\end{equation}
For simple factor graphs where there is no ambiguity, we may refer to factors
directly by the variables they link to. For instance, a simple fully-connected
factor graph with only two variables $\var{X}$ and $\var{Y}$ is fully specified
by 
$E(x, y) = E_\var{X}(x) + E_\var{Y}(y) + E_\var{XY}(x,y)\,.$

The energy function in \cref{eq:energy} induces preferences for certain
configurations. For instance, a globally best configuration can be found by
solving
$\argmin_{x,y} E(x,y)$,
while a best assignment for $\var{Y}$ given a fixed value of $\var{X}$
can be found by solving $\argmin_{y} E(x,y)$.%
\footnote{In this work, we only consider deterministic inference in factor
graphs. Probabilistic models are a promising extension.} %
We may think of, or suggest using notation, that $\var{X}$ is an \emph{input}
and $\var{Y}$ is an \emph{output}. However, intrinsically, factor graphs are not
attached to a static notion of input and output, and instead can be used to
infer any subset of variables given any other subset.

In our proposed framework of UNNs, 
we define the computation performed by a neural network  
using a factor graph, where each variable is a representation vector (\eg, analogous to the output of a layer in a standard network.)
We design the factor energy functions depending on the type of each variable and the desired relationships between them. 
Inference is performed by minimizing the joint energy with respect to all unobserved variables (\ie, hidden and output values). 
For instance, to construct a supervised UNN, we may designate a particular variable as ``input'' $\var{X}$ and another as ``output'' $\var{Y}$, alongside several hidden variables $\var{H}_i$,  compute
\begin{equation}\label{eq:inference}
\hat{y} = \arg\min_{y} \min_{h_1, \ldots, h_n} E(x, h_1, \ldots, h_n, y)\,,
\end{equation}
and train by minimizing some loss $\ell(\hat{y}, y)$.
However, UNNs are not restricted to the supervised setting or to a single input and output, as we shall explore.

While this framework is very flexible, \cref{eq:inference} is a 
non-trivial optimization problem. Therefore, we focus on
a class of energy functions that renders inference easier: 
\begin{equation}\label{eq:multilinear}
\begin{aligned}
E_{\var{X}_i}(x_i) &= -\DP{b_{\var{X}_i}}{x_i} + \Reg_{\var{X}_i}(x_i)\,,\\
E_{f}(x_f) &= -\bigDP{W_f}{\bigotimes_{\var{X}_j \in f} x_j }\,, \\
\end{aligned}
\end{equation}
where each $\Reg_{\var{X}_i}$ is a strictly convex regularizer, 
$\otimes$ denotes the outer product, and $W_f$ is a parameter tensor
of matching dimension.
For pairwise factors $f = \{\var{X}, \var{Y}\}$, the factor energy 
is bilinear and can be written simply as $E_{\var{XY}}(x, y) = -x^\top W y$.
In factor graphs of the form given in \cref{eq:multilinear}, the energy is convex in each variable separately, and block-wise minimization has a closed-form expression involving the Fenchel conjugate of the regularizers. This suggests a block coordinate descent optimization strategy: given an order $\pi$, iteratively set:
\begin{equation}\label{eq:coordinate_descent}
x_{\pi_j} \leftarrow \argmin_{x_{\pi_j}}~E(x_1, \ldots, x_n)\,.
\end{equation}
This block coordinate descent algorithm is guaranteed to decrease energy at every iteration and, for energies as in \cref{eq:multilinear}, to converge to a Nash equilibrium \citep[Thm.~2.3]{xu2013block}; in addition, it is conveniently learning-rate free. 
For training, to tackle the bi-level optimization problem, we unroll the coordinate descent iterations, 
and minimize some loss with standard deep learning optimizers, like stochastic gradient or Adam \citep{kingma2014adam}.




The following result, proved in \cref{sec:proof_lemma_unrolling}, shows that the coordinate descent algorithm (\cref{eq:coordinate_descent}) for UNNs with multilinear factor energies (\cref{eq:multilinear}), corresponds to standard forward propagation on an unrolled neural network.  
\begin{lemma}[Unrolling Lemma]\label{lemma:unrolling}
Let $\mathcal{G} = (V,F)$ be a pairwise factor graph, with multilinear higher-order energies and strictly convex unary energies, as in \cref{eq:multilinear}. 
Then, the coordinate descent updates \eqref{eq:coordinate_descent} result in a chain of affine transformations (\textit{i.e.}, pre-activations) followed by non-linear activations, applied in the order $\pi$, yielding a traditional computation graph.
\end{lemma}



We show next an undirected construction inspired by (directed) multi-layer perceptrons.



\paragraph{Single pairwise factor}



\begin{figure*}[ht]
    \centering
    \includegraphics[width=\textwidth]{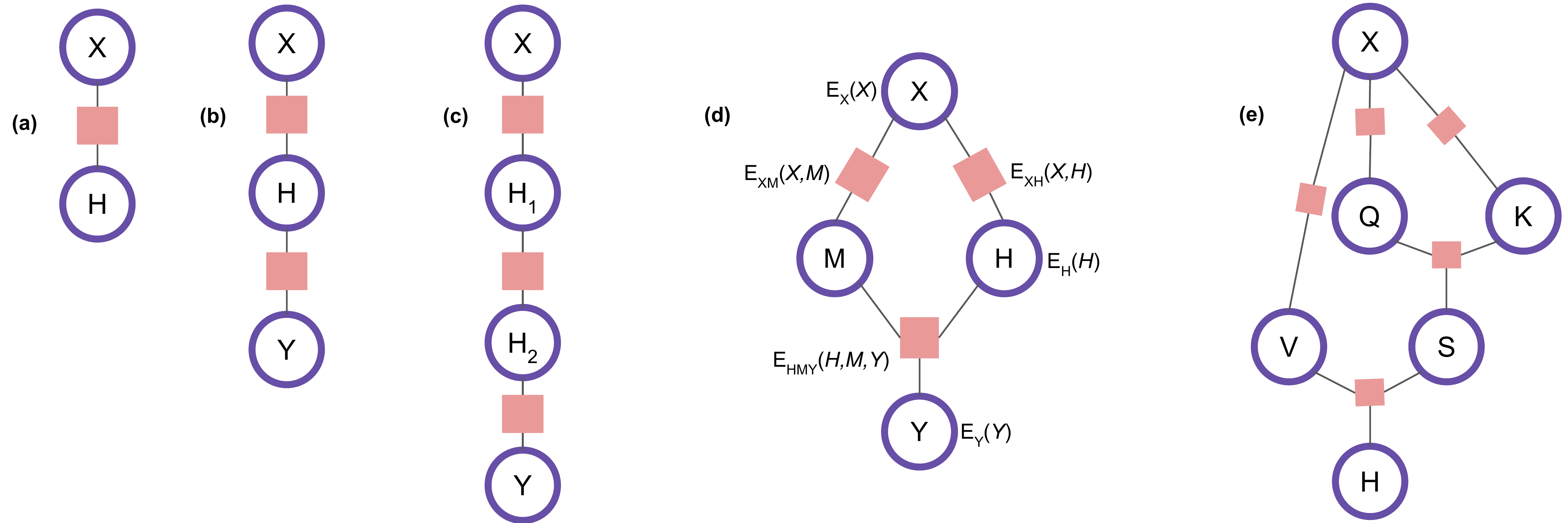}
    \caption{Factor graphs for:
    (a) network without intermediate layers,
    (b-c) undirected MLPs with one or two layers,
    (d) undirected biaffine dependency parser,
    (e) undirected self-attention.
Energy labels ommitted for brevity with the exception of (d).
    }
    \label{fig:factors_all_models}
\end{figure*}

The simplest possible UNN has a pairwise factor connecting two variables $\var{X,H}$. 
We may interpret $\var{X}$ as an input, and
$\var{H}$ either as an output (in supervised learning) or a hidden representation in unsupervised learning (\cref{fig:factors_all_models}(a)). 
%
Bilinear-convex energies as in \cref{eq:multilinear} yield:
\begin{equation}\label{eq:energy_rbm}
\begin{aligned}
E_{\var{XH}}(x,h) &= -\DP{h}{Wx}\,,\\
E_\var{X}(x) &= -\DP{x}{b_\var{X}} + \Reg_\var{X}(x)\,,\\
E_\var{H}(h) &= -\DP{h}{b_\var{H}} + \Reg_\var{H}(h)\,.\\
\end{aligned}
\end{equation}
This resembles a Boltzmann machine with continuous variables \citep{smolensky1986information,hinton2007boltzmann,welling2004exponential}; 
however, in contrast to  Boltzmann machines, we do not
model joint probability distributions,
but instead use factor graphs as representations 
of deterministic computation, more akin to computation graphs.

Given ${x}$,
the updated $h$ minimizing the energy is:
\begin{equation}
\begin{aligned}
    {h}_\star &= \argmin_{h \in \reals^{M}} -(W\currit{x}+b_\var{H})^{\top}h + \Reg_\var{H}(h) \\
    &= (\nabla \Reg_\var{H}^{*})(Wx+b_\var{H}),
\end{aligned}
\end{equation}
where $\nabla \Reg_\var{H}^{*}$ is the gradient of the conjugate function of $\Reg_\var{H}$. 
Analogously, the update for $\var{X}$ given $\var{H}$ is:
\begin{equation}
    \begin{aligned}
    \nextit{x}
    &= (\nabla \Reg_\var{X}^{*})(W^\top\currit{h}+ b_\var{X})\,.
\end{aligned}
\end{equation}
Other than the connection to Boltzmann machines,
one round of updates of $\var{H}$ and $\var{X}$ in this order
also describe the computation of an auto-encoder with shared encoder/decoder weights. %

\cref{table:unn_functions} shows examples of regularizers $\Reg$ and their corresponding $\nabla \Reg^{*}$,
In practice, we never evaluate $\Reg$ or $\Reg^*$, but only $\nabla \Reg^*$, 
which we choose among commonly-used neural network activation functions like $\tanh$, $\relu$, and $\softmax$.

\begin{table}[t]
    \renewcommand\arraystretch{1.5}%
    \centering%
    \caption{\label{table:unn_functions} Examples of regularizers $\Reg(h)$
corresponding to some common activation functions,
where $\phi(t) = t \log t.$}
    \vskip 0.15in
\begin{center}
\begin{small}
\begin{sc}
\begin{tabular}{ll}
    \toprule
    $\Reg(h)$ & $(\nabla \Reg^{*})(t)$  \\
    \midrule
    $\frac{1}{2} \| h \|^2$ & $t$ \\
    $\frac{1}{2}\| h \|^2 + \iota_{\reals_{+}}(h)$ & $\relu(t)$ \\
    $\sum_{j}\!\big(\phi(h_j) + \phi(1-h_j)\big) + \iota_{[0,1]^d}(h)$ &
$\operatorname{sigmoid}(t)$\\
    $\sum_{j}\!\left(\phi\left(\frac{1+h_j}{2}\right)
    +\phi\left(\frac{1-h_j}{2}\right)\right)
+ \iota_{[-1,1]^d}(h)$ & $\tanh(t)$ \\
    $-\entropy(h) + \iota_{\Delta}(h)$ & $\softmax(t)$ \\
    \bottomrule
\end{tabular}
\end{sc}
\end{small}
\end{center}
\end{table}

\begin{figure}[t]
    \centering
    \includegraphics[width=0.47\textwidth]{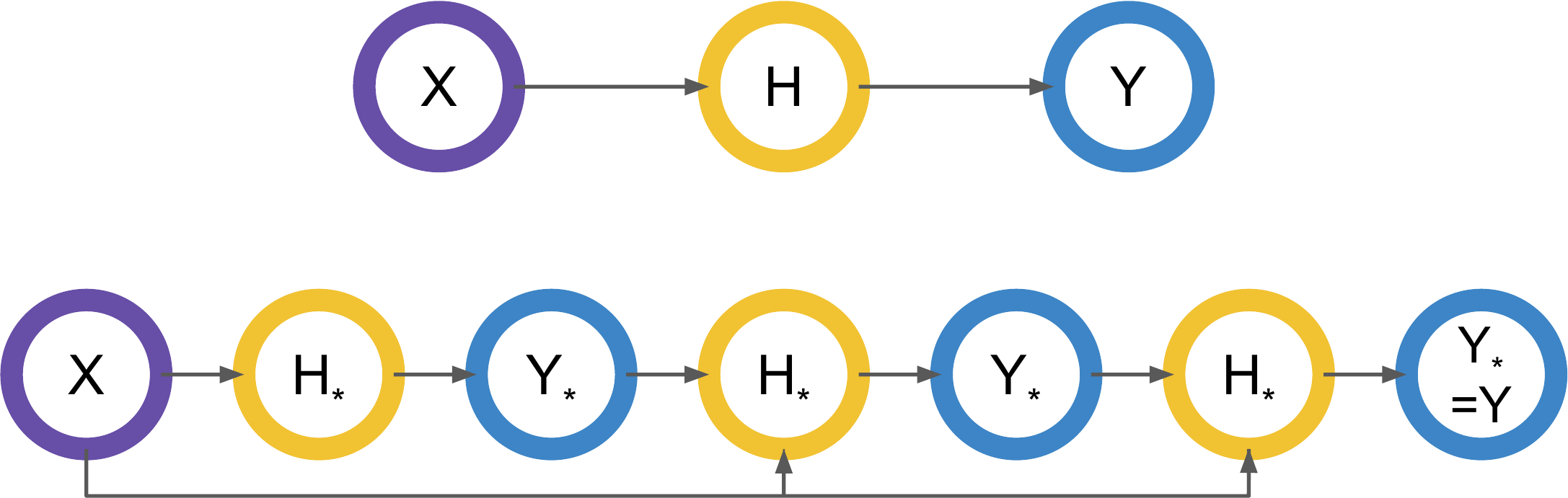}
    \caption{Unrolling the computation graph for undirected MLP with a single hidden layer. Top: MLP with one hidden layer. Bottom: Unrolled graph for UNN with $k=3$ iterations.}
    \label{fig:unroll_undirected_mlp}
\end{figure}

\paragraph{Undirected multi-layer perceptron (MLP)}
\cref{fig:factors_all_models}(b) shows the factor graph for an undirected MLP
analogous to a feed-forward one with input $\var{X}$, output $\var{Y}$,
and a single hidden layer $\var{H}$. 
As in \cref{eq:multilinear}, we have bilinear pairwise factors
\begin{equation}
    E_\var{XH}(x,h) = -\DP{h}{Wx}\,, \quad 
    E_\var{HY}(h,y) = -\DP{y}{Vh}\,,
\end{equation}
and linear-plus-convex unaries
$E_\var{Z}(x) = -\DP{x}{b_\var{Z}} + \Reg_\var{Z}(x)$
for $\var{Z}\in\{\var{X},\var{H},\var{Y}\}$.
If $x$ is observed (fixed), coordinate-wise
inference updates take the form:
\begin{equation}
\begin{aligned}
    \nextit{h} 
    &= 
     (\nabla \Reg_\var{H}^{*})(Wx+V^{\top}\currit{y}+b_\var{H})\,, \\
    \nextit{y} 
    &= 
     (\nabla \Reg_\var{Y}^{*})(V{h}+b_\var{Y})\,.
\end{aligned}
\end{equation}
Note that $E_\var{X}$ does not change anything if $\var{X}$ is always observed.
The entire algorithm
can be unrolled into a directed computation graph, leading to a deep neural
network with shared parameters (\cref{fig:unroll_undirected_mlp}). 

The regularizers $\Reg_\var{H}$ and $\Reg_\var{Y}$ may be selected based on what
we want $\nabla\Reg^*$ to look like, and the constraints or domains of the
variables. For instance, if $\var{Y}$ is a multiclass classification output,
we may pick $\Reg_\var{Y}$ such that
$\nabla\Reg_\var{Y}^*$ be the softmax function,
and $\Reg_\var{H}$ to induce a relu nonlinearity.  
Initializing $\initit{y}=0$ and performing a single iteration of updating $\var{H}$ followed by $\var{Y}$ results in a standard MLP with a single hidden layer (see also \cref{fig:unroll_undirected_mlp}).
However, the UNN point of view lets us decrease energy further by performing multiple iterations,
as well as use the same model to infer any variables given any other ones, \textit{e.g.}, to predict $x$ from $y$ instead of $y$ from $x$.
We demonstrate this power in
\cref{section:experiments_mnist,section:experiments_attention,section:experiments_parsing}.

The above constructions provide a flexible framework for defining UNNs. However, UNNs are more general and cover more popular deep learning architectures. The following constructions illustrate some such connections.

\paragraph{Feed-forward neural networks}

Any directed computation graph associated with a neural network is a particular case of an UNN. We illustrate this for a simple feed-forward network, which chains the functions $h = f(x)$ and $y = g(h)$, where $x \in \reals^m$, $h \in \reals^d$, $y \in \reals^n$ are input, hidden, and output variables, and $f: \reals^m \rightarrow \reals^d$ and $g: \reals^d \rightarrow \reals^n$ are the functions associated to each layer (\textit{e.g.}, an affine transformation followed by a non-linearity). This factor graph is illustrated in \cref{fig:factors_all_models}(b).
To see this, let $V = \{\var{X}, \var{H}, \var{Y}\}$ and $F = \{\var{XH},\var{HY}\}$ and define the energies as follows.
Let $d: \reals^d \times \reals^d \to \reals_+$ be any distance function satisfying $d(a,b) \ge 0$, with equality iff $a = b$; for example $d(a,b) = \|a-b\|$.
Let all the unary energies be zero and define the factor energies
$E_{\var{XH}}(x, h) = d(h,f(x))$ and $E_{\var{HY}}(h, y) = d(y,g(h))$. Then
the total energy satisfies $E(x, h, y) \ge 0$, with equality iff the
equations $h = f(x)$ and $y = g(h)$ are satisfied -- therefore, the energy is
minimized (and becomes zero) when $y=g(f(x))$, matching the corresponding directed computation graph.
This can be generalized for an arbitrary deterministic neural network. 
This way, we can form UNNs that are partly directed, partly undirected, as the whole is still an UNN. We do this in our experiments in \cref{section:experiments_parsing}, where we fine-tune a pretrained BERT model appended to a UNN for parsing.



\paragraph{Implicit layers}
UNNs include networks with implicit layers
\citep{duvenaud2020deep}, a paradigm which, in contrast with feed-forward
layers, does not specify how to compute the output from the input, but rather
specifies conditions that the output layer should specify, often related to
minimizing some function, \eg, computing a layer $h_{i+1}$ given a previous
layer $h_{i}$ involves solving a possibly difficult problem $\argmin_{h} f(h_i, h)$.
Such a function $f$ can be directly interpreted as an energy in our model, \ie,
$E_{\var{H}_i\var{H}_{i+1}} = f$.

%% file: experiments_mnist.tex
\section{Image Classification and Visualization}\label{section:experiments_mnist}

Unlike feed-forward networks, where the processing order is hard-coded from
inputs $\var{X}$ to outputs $\var{Y}$, UNNs support processing in any
direction. We can thus use the same trained network both for classification as well as for generating prototypical examples for each class.
We demonstrate this on the MNIST dataset of handwritten digits \citep{deng2012mnist},
showcasing convolutional UNN layers.


The architecture is shown in \cref{fig:factors_all_models}(c)
and has the following variables: the image $\var{X}$, the class label
$\var{Y}$ and two hidden layer variables $\var{H}_{\{1,2\}}$.
Unlike the previous examples, the two pairwise energies involving the image and
the hidden layers are convolutional, \ie, linear layers with internal structure:
\begin{equation}
\begin{aligned}
E_\var{XH_1}(X, H_1) &= -\DP{H_1}{\mathcal{C}_1(X; W_1)} \,, \\
E_\var{H_1H_2}(H_1, H_2) &= -\DP{H_2}{\mathcal{C}_2(H_1; W_2)}\,,
\end{aligned}
\end{equation}
where $\mathcal{C}_{1,2}$ are linear cross-correlation operators with
stride two and
filter weights $W_1 \in \reals^{32 \times 1 \times 6 \times 6}$
and
$W_2 \in \reals^{64 \times 32 \times 4 \times 4}$.
The last layer is fully connected:
\begin{equation}
E_\var{H_2Y}(H_2, y) = -\DP{V}{~y~\otimes H_2}\,.
\end{equation}
The unary energies for the hidden layers contain standard (convolutional) 
bias term along with the binary entropy term $\Psi_\text{tanh}(H)$ such that $\nabla
\Psi_\text{tanh}^*(t)=\operatorname{tanh}(t)$ (see \cref{table:unn_functions}).
Note that $\var{X}$ is no longer a constant when generating $\var{X}$ given $\var{Y}$, therefore it is important to specify the unary energy $E_\var{X}$.
Since pixel values are bounded, we set $\var{E}_X(x) = \Psi_\text{tanh}(x)$.
Initializing $H_1,H_2,$ and $y$ with zeroes and updating them once blockwise in
this order yields exactly a feed-forward convolutional neural network.
As our network is undirected, we may propagate information in multiple passes,
proceeding in the order $\var{H_1, H_2, Y, H_2}$ iteratively.
The
update for $H_1$ involves a convolution of $X$ and a deconvolution
of $H_2$; we defer the other updates to
\cref{section:conv_derivation}:
\begin{equation}
\nextit{(H_1)}\!\!\!\!\!=\operatorname{tanh}(\mathcal{C}_1(\currit{X}; W_1) +
\mathcal{C}^\top_2(\currit{H_2}; W_2) +
b_1 \otimes 1_{d_1})\,,
\end{equation}
where $b_1 \in \reals^{32}$ are biases for each filter,
and $d_1 = 12 \times 12$ is the convolved image size.
To generate digit prototype $X$ from a given class $c\in\{1,\ldots,10\}$, we may
set $y=e_c$, initialize the other variables at zero (including $\var{X}$),
and solve $\hat{X}=\argmin_X \min_{H_{1,2}} E(X, H_1, H_2, y)$ by coordinate descent in the reverse order $\var{H_2},\var{H_1},\var{X},\var{H_1}$ iteratively.

We train our model jointly for both tasks. For each labeled pair $(X,y)$ from
the training data, we first predict $\hat{y}$ given $X$, then separately predict $\hat{X}$
given $y$.
The incurred loss is a weighted combination
$\ell(x,y) = \ell_f(y, \hat{y}) + \gamma \ell_b(X, \hat{X})$,
where $\ell_f$ is a 10-class cross-entropy loss,
and $\ell_b$ is a binary cross-entropy loss averaged over all $28\times28$
pixels of the image.
We use $\gamma=.1$ and an Adam learning rate of $.0005$.

\begin{table}[t]
    \caption{MNIST accuracy with convolutional UNN.}
    \vskip 0.15in
\begin{center}
\begin{small}
\begin{sc}
    \begin{tabular}{ll}
    \toprule
        Iterations & Accuracy \\
        \midrule
        $k=1,\gamma=0$ (baseline) & 98.80 \\
        $k=1$ & 98.75 \\
        $k=2$ & 98.74 \\
        $k=3$ & \textbf{98.83} \\
        $k=4$ & 98.78 \\
        $k=5$ & 98.69 \\
    \bottomrule
    \end{tabular}
\end{sc}
\end{small}
\end{center}
    \label{table:mnist_results}
\end{table}





\begin{figure}[t]
    \def\figh{9cm}%
    \centering
    \subcaptionbox{}{
    \includegraphics[height=\figh]{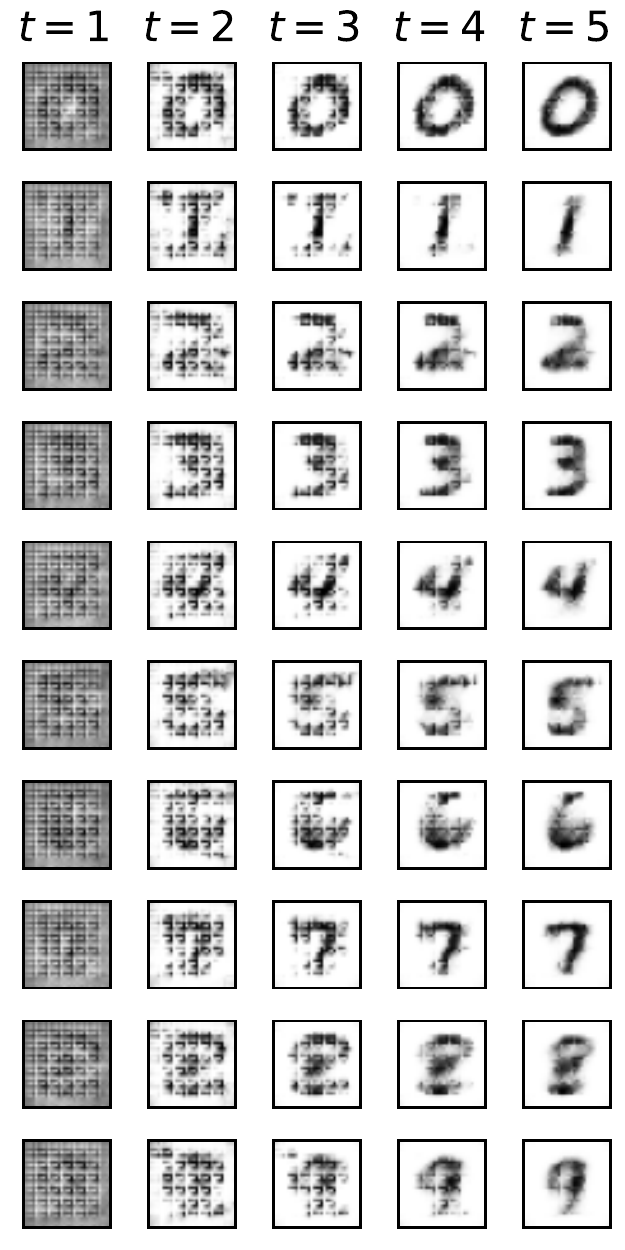}}\quad%
    \subcaptionbox{}{\includegraphics[height=\figh]{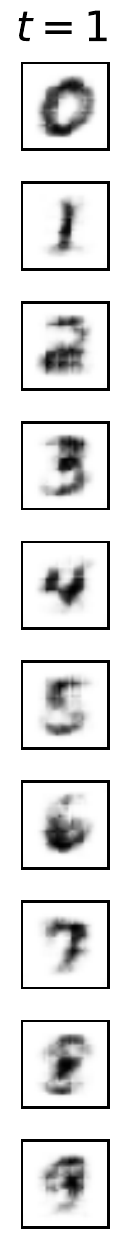}}\quad%
    \subcaptionbox{}{\includegraphics[height=\figh]{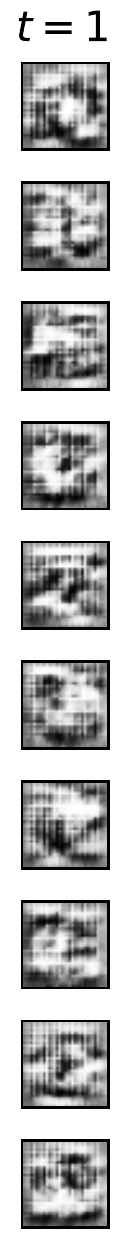}}%
    \caption{Digit prototypes generated by convolutional UNN.
(a) best UNN ($k=5,\gamma=.1)$, (b)
single iteration UNN ($k=1,\gamma=.1)$,
(c): standard convnet ($k=1,\gamma=0$).
\label{fig:mnist_generated}}
\end{figure}

The classification results are shown in \cref{table:mnist_results}. The model is able to achieve high classification accuracy, and multiple iterations lead to a slight improvement. This result suggests
the reconstruction loss for $\var{X}$ can also be seen as a regularizer,
as the same model weights are used in both directions.
The more interesting impact of multiple energy updates is the image prototype generation.
In \cref{fig:mnist_generated} we show the generated digit prototypes after several iterations of energy minimization, as well as for models with a single iteration.
The networks trained as UNNs produce recognizable digits, and in particular the model with more iterations learns to use the additional computation to produce clearer pictures.
As for the baseline, we may interpret it as an UNN and apply the same process to extract prototypes, but this does not result in meaningful digits (\cref{fig:mnist_generated}c). 
Note that our model is not a generative model – in that experiment, we are not sampling an image according to a probability distribution, rather we are using energy minimization deterministically to pick a prototype of a digit given its class.


\paragraph{Comparison of UNN to Unconstrained Model}
As per \cref{fig:unroll_undirected_mlp}, an unrolled UNN can be seen as a feed-forward network with a specific architecture and with weight tying. To confirm the benefit of the UNN framework, we compare against an unconstrained model, \ie, with the same architecture but separate, untied weights for each unrolled layer.
We use as a base the model described in forward-only mode and train a model with 2 to 5 layers with different weights instead of shared weights as in the case with the UNN. Depending on the number of layers, we cut the number of parameters in each layer, in order to obtain models with the same number of parameters as the UNN for fair comparison. The results from the experiment are described in \cref{tab:unconstrained}.

\begin{table}[ht]
    \centering
    \small
    \caption{Comparison of UNN with an unconstrained model with the same number of layers as the UNN iterations. The number of parameters of the UNN and the unconstrained model are roughtly the same.}
    \vskip 0.15in
    \begin{sc}
    \begin{tabular}{cccccc}
        \toprule
          & $k=1$ & $k=2$ & $k=3$ & $k=4$ & $k=5$ \\
         \midrule
         Acc. & 98.80 & 98.76 & 98.45 & 98.32 & 97.39 \\
         \midrule
         \# params & 50026 & 51220 & 51651 & 53608 & 51750 \\
        \bottomrule
    \end{tabular}
    \end{sc}
    \label{tab:unconstrained}
\end{table}

%% file: experiments_attention.tex
\section{Undirected Attention Mechanism}\label{section:experiments_attention}

Attention \citep{bahdanau2014neural,vaswani2017attention} is a key component that enables neural networks to handle variable-length sequences as input.
In this section, we propose an undirected attention mechanism
(\cref{fig:factors_all_models}(e)). We demonstrate this model on the task of completing missing values in a sequence of dynamic length $n$,
with the variable $\var{X}$ serving as both input and output, taking values $X \in
\reals^{d \times n}$, queries, keys and values taking values $Q, K, V \in
\reals^{n \times d}$, and attention weights $S \in (\triangle_n)^n$,
where $d$ is a fixed hidden layer size.
Finally, $\var{H}$ is an induced latent sequence representation, with values $H
\in \reals^{n \times d}$. The only trainable parameters are $W_\var{Q},
W_\var{K}, W_\var{V} \in \reals^{d \times d}$, and the input embeddings.
We model scaled dot-product attention given with
$\softmax(d^{-\frac{1}{2}}QK^\top)V$.
For all variables except $\var{S}$, we set $\energyltwo{\cdot}$. For the attention weights, we use $E_\var{S}(S) = -\sqrt{d}\sum_{i=1}^n \entropy(S_i)$.
The higher-order energies are:
\begin{equation}
\begin{aligned}
    E_\var{XQ}(X,Q) &= -\DP{Q}{W_\var{Q}(X+P)}\,,\\
    E_\var{XK}(X,K) &= -\DP{K}{W_\var{K}(X+P)}\,,\\
    E_\var{XV}(X,V) &= -\DP{V}{W_\var{V}(X+P)}\,,\\
    E_\var{QKS}(Q,K,S) &= -\DP{S}{Q K^\top}\,, 
    \\
    E_\var{VSH}(V,S,H) &= -\DP{H}{S V}\,,
\end{aligned}
\end{equation}
where $P$ is a matrix of sine and cosine positional embeddings of same dimensions as $X$ \cite{vaswani2017attention}.

Minimizing the energy yields the blockwise updates:
\begin{equation}
\begin{aligned}
    \nextit{Q} &= W_\var{Z}(\currit{X} + P) + S\currit{K}\,, \\
    \nextit{K} &= W_\var{K}(\currit{X} + P) + S^\top {Q}\,, \\
    \nextit{V} &= W_\var{V}(\currit{X} + P) + S^\top \currit{H}\,, \\
    \nextit{S} &= \softmax\left(d^{-1/2}({Q}{K}^\top + {V}{H}^{\top})\right)\,, \\
    \nextit{H} &= {S}{V}\,,\\
    \nextit{\bar{X}} &= {\bar{V}} W_\var{V} + {\bar{Q}} W_\var{Q} +
{\bar{K}} W_\var{K},
\end{aligned}
\end{equation}
where $\bar{X}$ denotes only the rows of $X$ corresponding to the masked
(missing) entries.

\begin{figure}[t]
    \centering
    \includegraphics[width=0.45\textwidth]{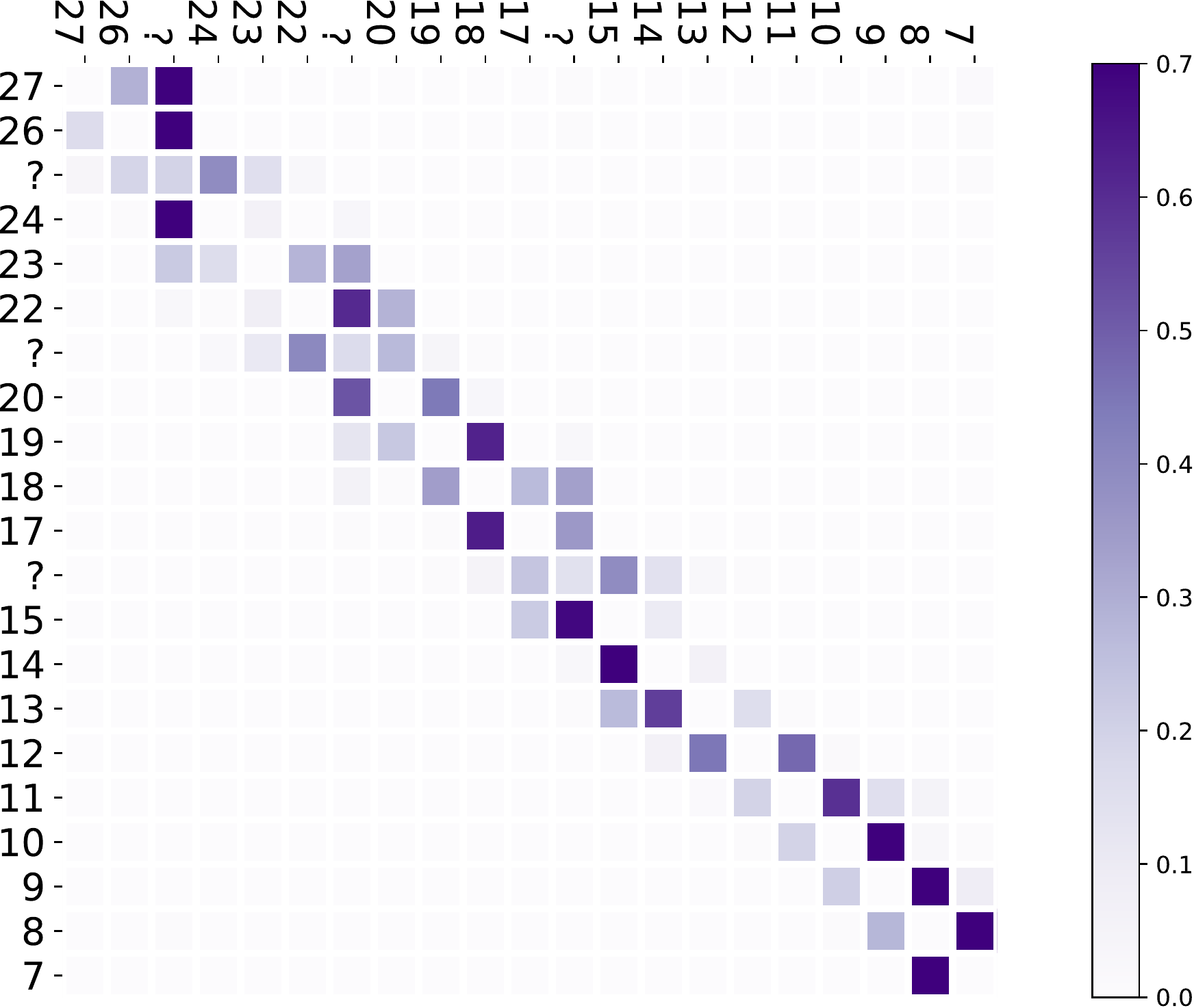}
    \caption{Undirected self-attention (one ``forward-backward'' pass) identifies an off-diagonal pattern, allowing generalization.
    }
    \label{fig:attn_example_small}
\end{figure}

Provided zero initialization, updating in the order $\var{(V/Q/K),S,H}$
corresponds exactly to a forward pass in a standard self-attention.
However, in an UNN, our expressions allow backward propagation back toward $\var{X}$, as well as iterating to an equilibrium.
To ensure that one round of updates propagates information through all the variables, we employ the ``forward-backward'' order
$\var{Q, K, V, S, H, S, V, K, Q, \bar{X}}$.



We evaluate the performance of the undirected attention with a toy task of sequence completion. We generate a toy dataset of numerical sequences between $1$ and $64$, of length at least 8 and at most 25, in either ascending or descending consecutive order. We mask out up to 10\% of the tokens and generate all possible sequences, splitting them into training and test sets with around 706K and 78K instances. %
Undirected self-attention is applied to the input sequence.
Note that because of the flexibility of the architecture, the update of the input variable $X$ does not differ from the updates of the remaining variables,
because each variable update corresponds to one step of coordinate descent.
The model incurs a cross-entropy loss for the missing elements of the sequence and the parameters are updated using Adam with learning rate $10^{-4}$. The hidden dimension is $d=256$, and
gradients with magnitude beyond $10$ are clipped.

Undirected attention is able to solve this task, reaching over 99.8\% test accuracy, confirming 
viablity. \Cref{fig:attn_example_small} shows the attention weights for a model trained with $k=1$. More attention plots are shown in \cref{appendix:additional_viz}.

\begin{figure}
    \centering
    \includegraphics[width=.99\linewidth]{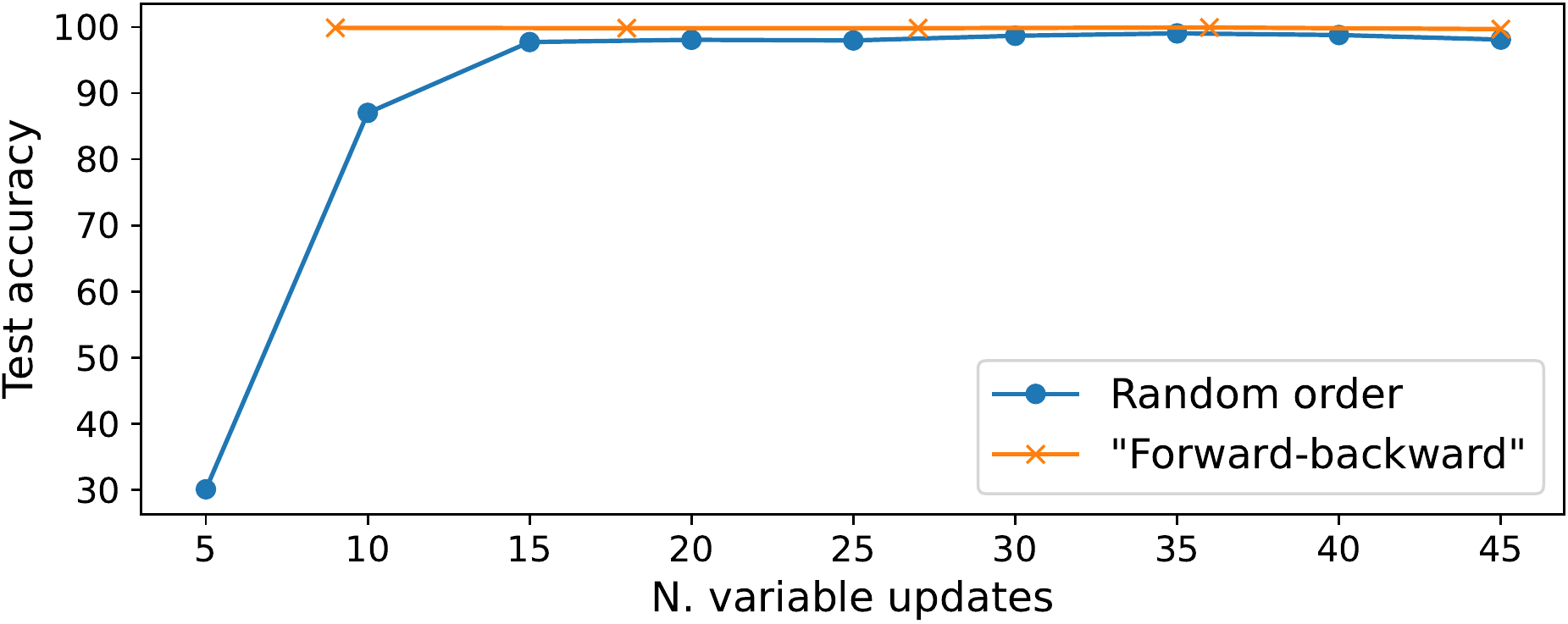}
    \caption{Comparison of the test accuracy) for models
with random and ``forward-backward'' order of variable updates. 
Markers indicate one full iteration.
}
\label{fig:attn_update_order_default_vs_random}
\end{figure}

\paragraph{Impact of update order.}
The ``forward-backward'' order is not the only possible order of updates in an UNN.
In \cref{fig:attn_update_order_default_vs_random} we compare its performance against a randomized coordinate descent strategy, where at each round we pick a permutation of $\var{Q,K,V,S,H}$. In both cases, $\var{\bar{X}}$ is updated at the end of each iteration.
(Note that the ``forward-backward'' order performs almost twice the number of updates per iteration.)
While the ``forward-backward'' order performs well even with a single pass, given sufficient iterations, random order updates can reach the same performance, suggesting
that hand-crafting a meaningful order helps but is not necessary.
Further analysis is reported in \cref{appendix:order_number_updates}.

%% file: experiments_parsing.tex
\section{Structured UNNs for Dependency Parsing}\label{section:experiments_parsing}

The concept of UNN can be applied to structured tasks -- all we need to do is to define \textit{structured factors}, as shown next.

\begin{figure*}[t]
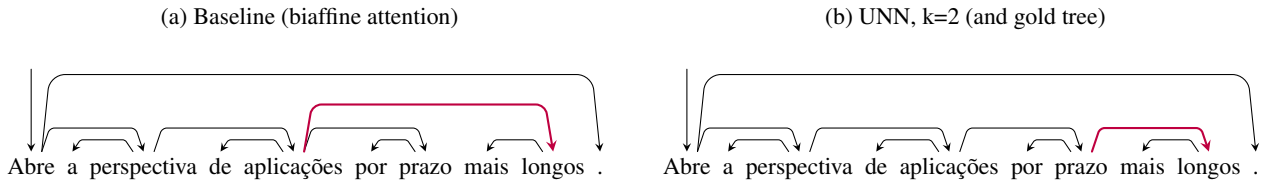

\centering\small
\begin{subfigure}[b]{.49\textwidth}
\caption{Baseline (biaffine attention)}
\begin{dependency}[hide label,edge unit distance=1.1ex]
\begin{deptext}[column sep=0.02cm]
Abre \&
a \&
perspectiva \&
de \&
aplicações \&
por \&
prazo \&
mais \&
longos \&
. \\
\end{deptext}
\depedge{1}{3}{1.0}
\depedge[edge unit distance=.8ex]{1}{10}{1.0}
\depedge{3}{2}{1.0}
\depedge{3}{5}{1.0}
\depedge{5}{4}{1.0}
\depedge{5}{7}{1.0}
\depedge[color=purple,thick]{5}{9}{1.0}
\depedge{7}{6}{1.0}
\depedge{9}{8}{1.0}
\deproot[edge unit distance=2.5ex]{1}{}
\end{dependency}
\end{subfigure}\quad
%
%
\begin{subfigure}[b]{.49\textwidth}
\caption{UNN, k=2 (and gold tree)}
\begin{dependency}[hide label,edge unit distance=1.1ex]
\begin{deptext}[column sep=0.02cm]
Abre \&
a \&
perspectiva \&
de \&
aplicações \&
por \&
prazo \&
mais \&
longos \&
. \\
\end{deptext}
\depedge{1}{3}{1.0}
\depedge[edge unit distance=.8ex]{1}{10}{1.0}
\depedge{3}{2}{1.0}
\depedge{3}{5}{1.0}
\depedge{5}{4}{1.0}
\depedge{5}{7}{1.0}
\depedge{7}{6}{1.0}
\depedge[color=purple,thick]{7}{9}{1.0}
\depedge{9}{8}{1.0}
\deproot[edge unit distance=2.5ex]{1}{}
\end{dependency}
\end{subfigure}
    \caption{Examples of dependency trees produced by the parsing model for a sentence in Portuguese. The baseline model (a)
    erroneously assigns the noun \textit{aplicações} as the syntactic head of the adjective \textit{longos}. The UNN with $k=2$ iterations (b) matches the gold parse tree for this sentence, eventually benefiting from the structural information propagated back from the node $\var{Y}$ after the first iteration.}
    \label{fig:example_dependency_tree}
\end{figure*}

We experiment with a challenging structured prediction task from natural language processing: unlabeled, non-projective dependency parsing \citep{kubler2009dependency}.
Given a sentence with $n$ words, represented as a matrix $X \in \reals^{r \times n}$ (where $r$ is the embedding size), the goal is to predict the syntactic relations as a \textit{dependency tree}, \textit{i.e.}, a spanning tree which has the words as nodes. The output can be represented as a binary matrix  $Y \in \reals^{n \times n}$, where the $(i,j)\textsuperscript{th}$ entry indicates if there is a directed arc $i \rightarrow j$ connecting the $i\textsuperscript{th}$ word (the \textit{head}) and $j\textsuperscript{th}$ word (the \textit{modifier}).
\cref{fig:example_dependency_tree} shows examples of dependency trees produced by this model.
We use a probabilistic model where the output $Y$ can more broadly represent a probability distribution over trees, represented by the matrix of arc marginals induced by this distribution (illustrated in \cref{appendix:additional_viz}.)

\paragraph{Biaffine parsing.}
A successful model for dependency parsing is the biaffine one
\citep{dozat2016deep, kiperwasser-goldberg-2016-simple}.
This model first computes head representations $H \in \reals^{d \times n}$ and modifier representations $M \in \reals^{d \times n}$,
via a neural network that takes $X$ as input -- here, $d$ denotes the hidden dimension of these representations.
Then, it computes a score matrix as $Z = H^\top V M \in \reals^{n \times n}$, where $V \in \reals^{d \times d}$ is a parameter matrix. 
Entries of $Z$  can be interpreted as scores for each candidate arc.
From $Z$, the 
best tree is found by the Chu-Liu-Edmonds  maximum spanning arborescence algorithm \citep{Chu1965, Edmonds1967}, and
probabilities and marginals by the matrix-tree theorem \citep{koo2007structured,smith2007probabilistic,mcdonald2007complexity,Kirchhoff1847}.

\paragraph{UNN for parsing.}
We now construct an UNN with the same building blocks as this biaffine  model, leading to the factor graph in \cref{fig:factors_all_models}(d).
The variable nodes are $\{\var{X}, \var{H}, \var{M}, \var{Y}\}$,
and the factors are $\{\var{XH}, \var{XM}, \var{HMY}\}$.
Given parameter weight matrices $V, W_\var{H}, W_\var{M} \in \reals^{d \times
d}$ and biases $b_\var{H}, b_\var{M} \in \reals^{d}$,
we use multilinear factor energies as follows: 
\begin{equation}
\begin{aligned}\label{eq:factor_energies_parsing}
    E_{\var{XH}}(X,H) &= -\DP{H}{W_\var{H} X}\,,\\
    E_{\var{XM}}(X,M) &= -\DP{M}{W_\var{M} X}\,, \\
    E_\var{YHM}(Y,H,M) &=  -\DP{Y}{H^\top VM}\,.
\end{aligned}
\end{equation}

For $\var{H}$ and $\var{M}$, we use the ReLU regularizer,
\begin{equation}\label{eq:unary_energies_parsing}
\begin{aligned}
    E_\var{H}(H)
    &= -\DP{b_\var{H} \otimes 1_n}{H} +
    \frac{1}{2}\|H\|^2 + \iota_{\geq 0}(H)\,\\
    E_\var{M}(M)
    &= -\DP{b_\var{M} \otimes 1_n}{M} +
    \frac{1}{2}\|M\|^2 + \iota_{\geq 0}(M)\,.\\
\end{aligned}
\end{equation}
For $\var{Y}$, however, we employ a structured entropy regularizer:
\begin{equation}
\begin{aligned}
    E_\var{Y}(Y) &= -\entropy_\mathcal{M}(Y) + \iota_{\mathcal{M}}(Y)\,,
\end{aligned}
\end{equation}
where $\mathcal{M}=\conv(\mathcal{Y})$ is the marginal polytope
\citep{wainwright2008graphical,martins2009polyhedral},
the convex hull of the adjacency matrices of all valid non-projective dependency trees (\cref{fig:example_marginals}), and $\entropy_\mathcal{M}(Y)$ is the maximal entropy over all distribution over trees with arc marginals $Y$:
\begin{equation}
\entropy_\mathcal{M}(Y) \coloneqq \max_{\alpha \in \triangle_{|\mathcal{Y}|}}
\mathcal{H}(\alpha) 
~\text{s.t.}~
\mathbb{E}_{A\sim\alpha}[A] = Y
\,.
\end{equation}

\paragraph{Derivation of block coordinate descent updates.}

To minimize the total energy,
we iterate between updating $H$, $M$ and $Y$ $k$ times, similar to the unstructured case. 

The updates for the heads and modifiers work out to: 
\begin{equation}
\begin{aligned}
\label{eq:derive_HM}
\nextit{H} &= \relu(W_\var{H} X +  b_\var{H} \otimes 1_n + V\currit{M}\currit{Y}^\top)\,,\\
\nextit{M} &= \relu(W_\var{M}X + b_\var{M} \otimes 1_n + V^\top {H}\currit{Y})\,.
\end{aligned}
\end{equation}
%
For $\var{Y}$, however, we must solve the problem
\begin{equation}
\begin{aligned}\label{eq:derive_Y}
    \nextit{Y} 
    &= \argmin_{Y \in \mathcal{M}} -\DP{Y}{{H}^\top V{M}}
-\entropy_\mathcal{M}(Y)\,.\\
\end{aligned}
\end{equation}
This combinatorial optimization problem corresponds to \emph{marginal 
inference} \citep{wainwright2008graphical}, 
a well-studied computational problem in structured prediction that appears in all probabilistic models. 
While generally intractable, for non-projective dependency
parsing it may be computed in time $\mathcal{O}(n^3)$ via
the aforementioned matrix-tree theorem, the same algorithm required to compute the structured likelihood loss.%
\footnote{During training, the matrix-tree theorem can be invoked only once to compute both the update to $Y$ as well as the gradient of the loss, since 
$\nabla \log p(\var{Y}=Y_\text{true}) = 
Y_\text{true}
- \hat{Y}$.}

With zero initialization, the first iteration yields the same hidden representations
and output as the biaffine model, assuming the updates are performed
in the order described.
The extra terms involving $VMY^\top$ and $V^\top H Y$ enable the current prediction for $Y$ to influence neighboring words, which leads to a more expressive model overall.



\begin{table}[t]
    \caption{Structured UNN parsing results.
    Columns show the number of UNN iterations.
    The best result per row is rendered in bold.
    }
    \vskip 0.15in
\begin{center}
\begin{small}
\begin{sc}
    \begin{tabular}{*6c}
    \toprule
    Language & $k=1$ & $k=2$ & $k=3$ & $k=4$ & $k=5$ \\
    \midrule 
    \multicolumn{6}{c}{Unlabeled attachment score} \\
    \midrule
    AF & \textbf{89.09} & 88.98 & 88.40 & 87.77 & 88.46 \\
    AR & \textbf{85.62} & 84.94 & 84.22 & 83.69 & 83.63 \\
    CS & 93.79 & \textbf{93.83} & 93.82 & 93.60 & 93.77 \\
    EN & 91.96 & 91.86 & 91.09 & \textbf{91.99} & 91.51 \\
    FA & \textbf{83.41} & 83.27 & 82.95 & 83.37 & 83.27 \\
    HU & 85.11 & \textbf{85.77} & 84.47 & 85.13 & 84.09 \\
    IT & \textbf{94.76} & 94.43 & 94.35 & 94.59 & 94.45 \\
    PT & 96.99 & 97.00 & 96.83 & \textbf{97.06} & 96.90 \\
    SW & \textbf{91.42} & 90.92 & 91.30 & 91.08 & 90.98 \\
    TE & 89.72 & 89.72 & \textbf{90.00} & 88.45 & 87.75 \\

    \midrule
    \multicolumn{6}{c}{Modifier list accuracy} \\
    \midrule
    AF &  \textbf{74.10} & 72.60 & 72.90 & 71.78 & 72.01 \\
    AR &  \textbf{70.44} & 69.29 & 68.41 & 68.08 & 68.19 \\
    CS &  84.46 & 84.82 & \textbf{84.93} & 84.12 & 84.49 \\
    EN &  79.08 & 77.73 & 75.20 & 78.90 & \textbf{79.44} \\
    FA &  64.80 & \textbf{66.75} & 65.28 & 66.67 & 65.85 \\
    HU &  64.13 & \textbf{66.07} & 64.37 & 62.91 & 64.13 \\
    IT &  \textbf{85.32} & 83.59 & 83.71 & 83.94 & 84.05 \\
    PT &  90.10 & \textbf{90.69} & 90.39 & 90.66 & 90.49 \\
    SW &  \textbf{79.07} & 78.37 & 78.52 & 78.60 & 78.24 \\
    TE &  72.87 & 72.87 & \textbf{73.68} & 66.80 & 65.99 \\

    \midrule
    \multicolumn{6}{c}{Exact match} \\
    \midrule
    AF &  \textbf{37.70} & 33.88 & 34.43 & 33.88 & 32.79 \\ 
    AR &  19.44 & 19.29 & 18.36 & \textbf{19.91} & 18.36 \\
    CS &  59.17 & 60.76 & \textbf{60.92} & 59.42 & 59.84 \\
    EN &  \textbf{48.59} & 44.37 & 40.14 & 43.66 & 44.37 \\
    FA &  21.52 & 22.15 & 22.78 & \textbf{24.68} & 23.42 \\
    HU &  21.13 & 23.40 & \textbf{24.15} & 23.40 & 21.51 \\
    IT &  \textbf{64.93} & 63.54 & 62.85 & 63.89 & 64.24 \\
    PT &  73.24 & \textbf{74.86} & 73.89 & 74.43 & 74.11 \\
    SW &  \textbf{54.62} & 52.38 & 54.13 & 53.94 & 52.67 \\
    TE &  75.69 & 77.08 & \textbf{79.17} & 71.53 & 70.14 \\
    \bottomrule
    \end{tabular}
    \end{sc}
    \end{small}
    \end{center}
    \label{table:parsing_results}
\end{table}

\paragraph{Experiments.}

We test the architecture on several datasets from Universal Dependencies 2.7 \citep{universal-dependencies-2.7}, covering different language families and dataset size: Afrikaans (AfriBooms), Arabic (PADT), Czech (PDT), English (Partut), Hungarian (Szeged), Italian (ISDT), Persian (Seraji), Portuguese (Bosque), Swedish (Talbanken), and Telugu (MTG). Performance is measured by three metrics:
\begin{itemize}
    \item Unlabeled attachment score (UAS): a fine-grained, arc-level accuracy metric.
    \item Modifier list accuracy: the percentage of head words for which \textit{all} modifiers were correctly predicted. For example, in \cref{fig:example_dependency_tree}, the baseline correctly predicts all modifiers for the words \emph{perspectiva, abre, longos}, but not for the words \emph{aplicações, prazo}.
    \item \looseness=-1
    Exact match: the percentage of sentences for which the entire parse tree is correct; the harshest of the metrics.
\end{itemize}
The latter, coarser measures can give more information whether the model is able to learn global relations, not just how to make local predictions correctly (\textit{i.e.}, when only prediction of the arcs is evaluated).

Our architecture is as follows: First, we pass the sentence through a BERT model (\textsc{bert-base-multilingual-cased}, fine-tuned during training, as directed networks can be added as components to UNNs, as mentioned in \cref{section:unn}) 
and get the word representations of the last layer. These representations are
the input $x$ in the UNN model. Then, we apply the parsing model described in
this section. The baseline ($k=1$) corresponds to a biaffine parser using BERT features.
The learning rate for each language is chosen via grid search for highest UAS on the validation set for the baseline model. We searched over the values $\{0.1, 0.5, 1, 5, 10\} \times 10^{-5}$.
In the experiments, we use $10^{-5}$ for Italian and $5 \times 10^{-5}$ for the other languages. 
We employ dropout regularization, using the same dropout mask for each variable throughout the inner coordinate descent iterations, so that dropped values do not leak.

The results from the parsing experiments are displayed in \cref{table:parsing_results}.
The numbers in the table show results on the test set for the highest validation accuracy epoch.
We see that some of the languages seem to benefit from the iterative procedure of UNNs (CS, HU, TE), while others do not (EN, AF), and little difference is observed in the remaining languages. 
In general, the baseline ($k=1$) seems to attain higher accuracies in UAS (individual arcs), but most of the languages have overall more accurate structures (as measured by modifier list accuracy and by exact match)
for $k>1$. 
\cref{fig:example_dependency_tree} illustrates with one example in Portuguese. 


%% file: related_work.tex
\section{Related Work}\label{section:related_work}



Besides the models mentioned in \cref{section:unn}  which may be regarded as particular cases of UNNs, other models and architectures, next described, bear relation to our work. 


\paragraph{Probabilistic modeling of joint distributions}

Our work draws inspiration from the well-known Boltzmann machines and Hopfield networks \citep{ackley1985learning,smolensky1986information,hopfield1984neurons}. 
We consider deterministic networks whose desired configurations minimize an energy function which decomposes as a factor graph. In contrast, 
many other works have studied probabilistic energy-based models (EBM) defined as Gibbs distributions, as well as efficient methods to learn those distributions and to sample from them \citep{ngiam2011learning,du_mordatch}. 
Similar to how our convolutional UNN can be used for multiple purposes
in \cref{section:experiments_mnist}, 
\citet{Grathwohl2020Your} reinterprets standard discriminative classifiers $p(\var{Y}|\var{X})$ as an EBM of a joint distribution $p(\var{X},\var{Y})$. 
Training stochastic EBMs requires Monte Carlo sampling or auxiliary 
networks \citep{grathwohl2020learning}; in contrast,
our deterministic UNNs, more aligned conceptually with deterministic EBMs \citep{lecun2006tutorial}, eschew probabilistic modeling in favor of
more direct training.
Moreover, our UNN architectures closely parallel feed-forward networks and reuse their building blocks, uniquely bridging the two paradigms.

\paragraph{Structured Prediction Energy Networks (SPENs)}
We saw in \cref{section:experiments_parsing} that UNNs can handle structured outputs. An alternative framework for expressive structured prediction is given by SPENs \citep{belanger2016structured}. 
Most SPEN inference strategies require gradient descent, often with higher-order gradients for learning \citep{belanger2017end}, or training separate inference networks \citep{tu-etal-2020-improving}.
UNNs in contrast, are well suited for coordinate descent
inference: a learning-rate free algorithm with 
updates based on existing neural network building blocks.
An undirected variant of SPENs would be similar to the MLP factor graph in \cref{fig:factors_all_models}(b), but with $\var{X}$ and $\var{Y}$ connected to a joint, higher-order factor, rather than via a
chain $\var{X-H-Y}$.

\paragraph{Universal transformers and Hopfield networks}
In \cref{section:experiments_attention} we show how we can implement self-attention with UNNs. Performing multiple energy updates resembles -- but is different from -- transformers \citep{vaswani2017attention} with shared weights between the layers. 
Our perspective of minimizing UNNs with coordinate descent using a fixed schedule and this unrolling is similar (but not exactly the same due to the skip connections) to having deeper neural networks which shared parameters for each layer. Such an architecture is the Universal Transformer \citep{dehghani2018universal}, which applies a recurrent neural network to the transformer encoder and decoder.
Recent work \citep{ramsauer2020hopfield} shows that the self-attention layers of transformers can be regarded as the update rule of a Hopfield network with continuous states \citep{hopfield1984neurons}. This leads to a ``modern Hopfield network'' with continuous states and an update rule which ensures global convergence to stationary points of the energy (local minima or saddle points). 
Like that model, UNNs also seek local minima of an energy function, albeit with a different goal. 

\paragraph{Deep models as graphical model inference.}
This line of work defines neural computation via approximate inference in
graphical models.
\citet{domke2012optimizationbasedlearning} derives backpropagating versions of gradient descent, heavy-ball and LBFGS. They require as input only routines to compute the gradient of the energy with respect to the domain and parameters. 
\citet{domke2012optimizationbasedlearning} studies learning with unrolled gradient-based inference in general energy models.
UNNs, in contrast, allow efficient, learning rate free, block-coordinate optimization by design.
An exciting line of work derives unrolled architectures
from inference in specific generative models
\citep{hershey2014deepunfolding,li2014meanfieldnetworks,lawson2019energyinspiredmodels}---a powerful construction at the cost of more
challenging optimization. 
The former is closest to our strategy, but by starting from probabilistic models the resulting updates are farther from
contemporary deep learning (\eg, convolutions, attention). In contrast, UNNs can reuse successful implementations, modular pretrained models, as well as structured factors, as we demonstrate in our parsing experiments.
We believe that our UNN construction
can shed new light over probabilistic inference models as well, uncovering deeper connections between the paradigms.

%% file: conclusions.tex
\section{Conclusions and Future Work}

We presented UNNs -- a structured energy-based model which combines the power of factor graphs and neural networks. At inference time, the model energy is minimized with a coordinate descent algorithm, allowing reuse of existing building blocks in a modular way with guarantees of decreasing the energy at each step. 
We showed how the proposed UNNs subsume many existing architectures, conveniently combining supervised and unsupervised/self-supervised learning, as demonstrated on the three tasks.

We hope our first steps in this work will spark multiple directions of future work on undirected networks. One promising direction is on probabilistic UNNs with Gibbs sampling, which have the potential to bring our modular architectures to  generative models.
Another direction is to consider alternate training strategies for UNNs.
Our strategy of converting UNNs to unrolled neural networks, enabled by \cref{lemma:unrolling}, 
makes gradient-based training easy  to implement,
but alternate training strategies, perhaps based on equilibrium conditions or dual decomposition, hold promise. 
Promising directions could be using this framework for dealing with missing data or learning the joint probability distribution.

%% file: acknowledgements.tex
\section*{Acknowledgements}
We would like to thank M\'ario Figueiredo, Caio Corro and the DeepSPIN team for helpful discussions. TM and AM are supported by the European Research Council (ERC StG DeepSPIN 758969)  
and by the Fundação para a Ciência e Tecnologia through contracts PTDC/CCI-INF/4703/2021 (PRELUNA) and  UIDB/50008/2020. VN is partially supported by the Hybrid Intelligence Centre, a 10-year programme funded by the Dutch Ministry of Education, Culture and Science through the Netherlands Organisation for Scientific Research (\href{https://hybrid-intelligence-centre.nl/}{https://hybrid-intelligence-centre.nl/}).

%% file: appendices.tex
\newpage
\appendix
\onecolumn

\section{Proof of \cref{lemma:unrolling}}\label{sec:proof_lemma_unrolling}

We provide a more general proof for multilinear factor potentials, of which  bilinear potentials are a special case. 
Let $\mathcal{G} = (V,F)$ be the factor graph underlying the UNN, with energy function $E(x_1, \ldots, x_n) = \sum_i E_{\var{X}_i}(x_i) + \sum_f E_{f}(x_f)$. We assume $E_{\var{X}_i}(x_i) = -b_i^\top x_i + \Psi_{\var{X}_i}(x_i)$ for each $\var{X}_i \in V$, with $\Psi_{\var{X}_i}$ convex, and $E_{f}(x_f) = -\DP{W_f}{\otimes_{j \in f} x_j}$ for each higher order factor $f \in F$ (multilinear factor energy), 
where $\otimes$ is the outer product, and $W_f$ is a parameter tensor
of matching dimension.
For pairwise factors $f = \{\var{X}_i, \var{X}_j\}$, the factor energy is bilinear and can be written simply as $E_{f}(x_i, x_j) = -x_i^\top W_f x_j$.

The (block) coordinate descent algorithm updates each representation $x_i \in V$ sequentially, leaving the remaining representations fixed. 
Let $F(\var{X}_i) = \{f \in F \,:\, \var{X}_i \in f\} \subseteq F$ denote the set of factors $\var{X}_i$ is linked to. 
The updates can be written as:
\begin{align}
    (x_i)_\star &= \arg\min_{x_i} E_{\var{X}_i}(x_i) + \sum_{f \in F(\var{X}_i)} E_f(x_f)\nonumber\\
    &= \arg\min_{x_i} \Psi_{\var{X}_i}(x_i) \underbrace{-b_i^\top x_i - \sum_{f \in F(\var{X}_i)} \DP{W_f}{\displaystyle\bigotimes_{j \in f} x_j}}_{-z_i^\top x_i}\nonumber\\
    &= (\nabla \Psi_{\var{X}_i}^*)(z_i),
\end{align}
where 
$z_i$ is a pre-activation given by 
\begin{align}
    z_i &= \left(\sum_{f \in F(\var{X}_i)} \rho_i(W_f) \bigotimes_{j \in f, j \ne i} x_j\right) + b_i,
\end{align}
and $\rho_i$ is the linear operator that reshapes and rolls the axis of $W_f$ corresponding to $x_i$ to the first position.  
If all factors are pairwise, the update is more simply:
\begin{align}
    (x_i)_\star &= (\nabla \Psi_{\var{X}_i}^*)\left(\sum_{f = \{\var{X}_i, \var{X}_j\} \in F(\var{X}_i)} \rho_i(W_f) x_j + b_i\right),
\end{align}
where $\rho_i$ is either the identity or the transpose operator.
The update thus always consists in applying a (generally non-linear) transformation $\nabla \Psi_{\var{X}_i}^*$ to an affine transformation of the neighbors of $\var{X}_i$ in the graph (that is, the variables that co-participate in some factor). 

Therefore, given any topological order of the variable nodes in $V$, running $k$ iterations of the coordinate descent algorithm following that topological order is equivalent to performing forward propagation in an (unrolled) directed acyclic graph, where each node applies affine transformations on input variables followed by the activation function $\nabla \Psi_{\var{X}_i}^*$.






\section{Analysis of Order and Number of Variable Updates}\label{appendix:order_number_updates}

For one of the experiments - undirected self-attention, we analyze how the order of variable updates and the number of update passes during training affect the model performance. 

\paragraph{Order of variable updates.} 
In \cref{section:experiments_attention} we showed that one pass of the ``forward-backward'' order or variable updates ($\var{Q,K,V,S,H,S,V,K,Q,\hat{X}}$) performs well enough for the of sequence completion. 
Since the flexibility of our model does not limit us to a specific order, we compare it to a random order of updating the variables (a permutation of $\var{Q,K,V,S,H}$; $\var{\hat{X}}$ is always updated last). One pass over the ``forward-backward'' order performs nine variable updates, and one pass over the random order - five. In \cref{fig:attn_update_order_default_vs_random} we show how the two ways of order perform for different number of variable updates (for example, 2 passes over the ``forward-backward'' model equal 18 variable updates, and over the random model - 10). 
The ``forward-backward'' order performs best, but the random order can achieve similar performance after enough number of updates.

\begin{figure}[ht!]
    \centering
    \includegraphics[width=.45\textwidth]{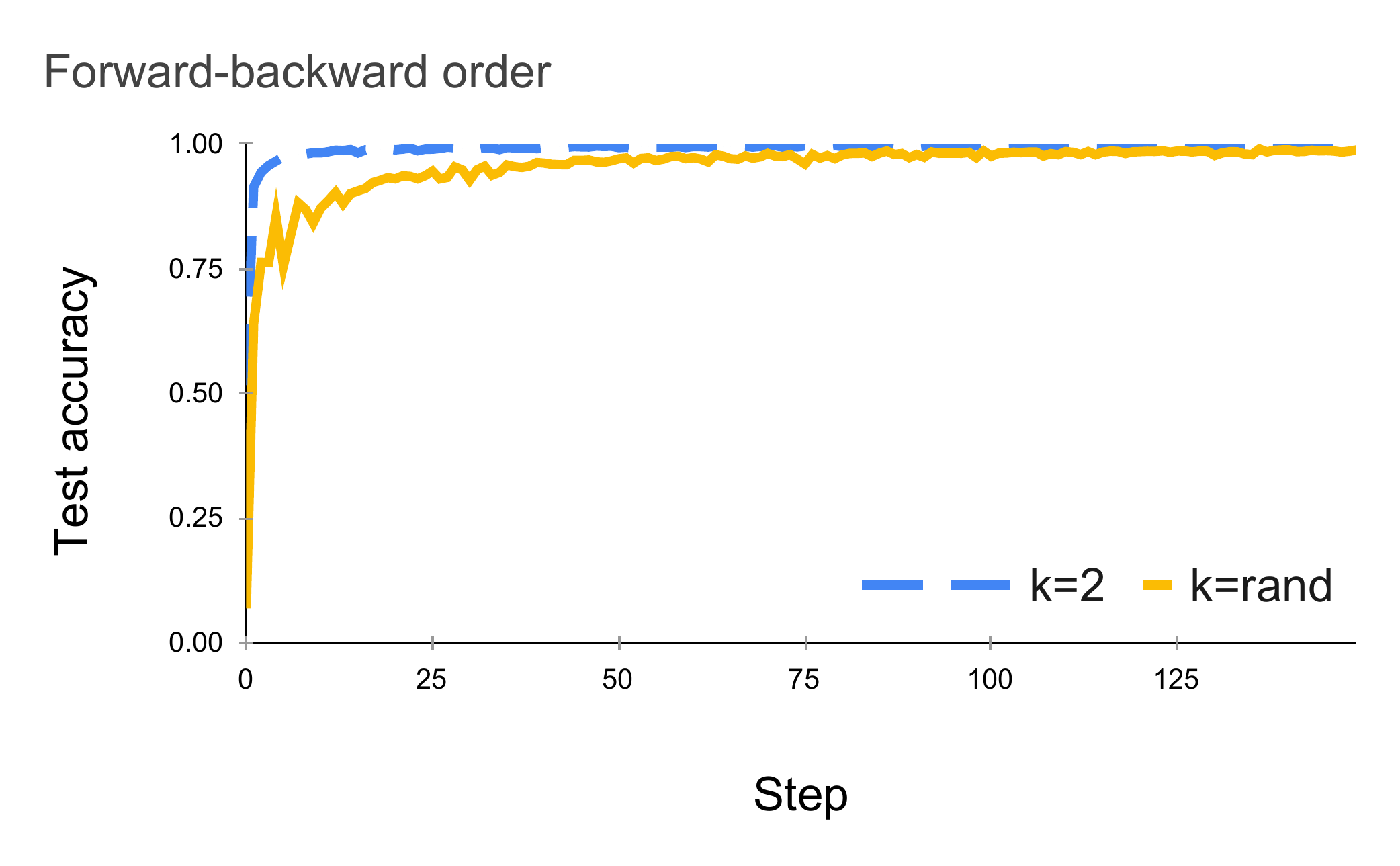}
    \includegraphics[width=.45\textwidth]{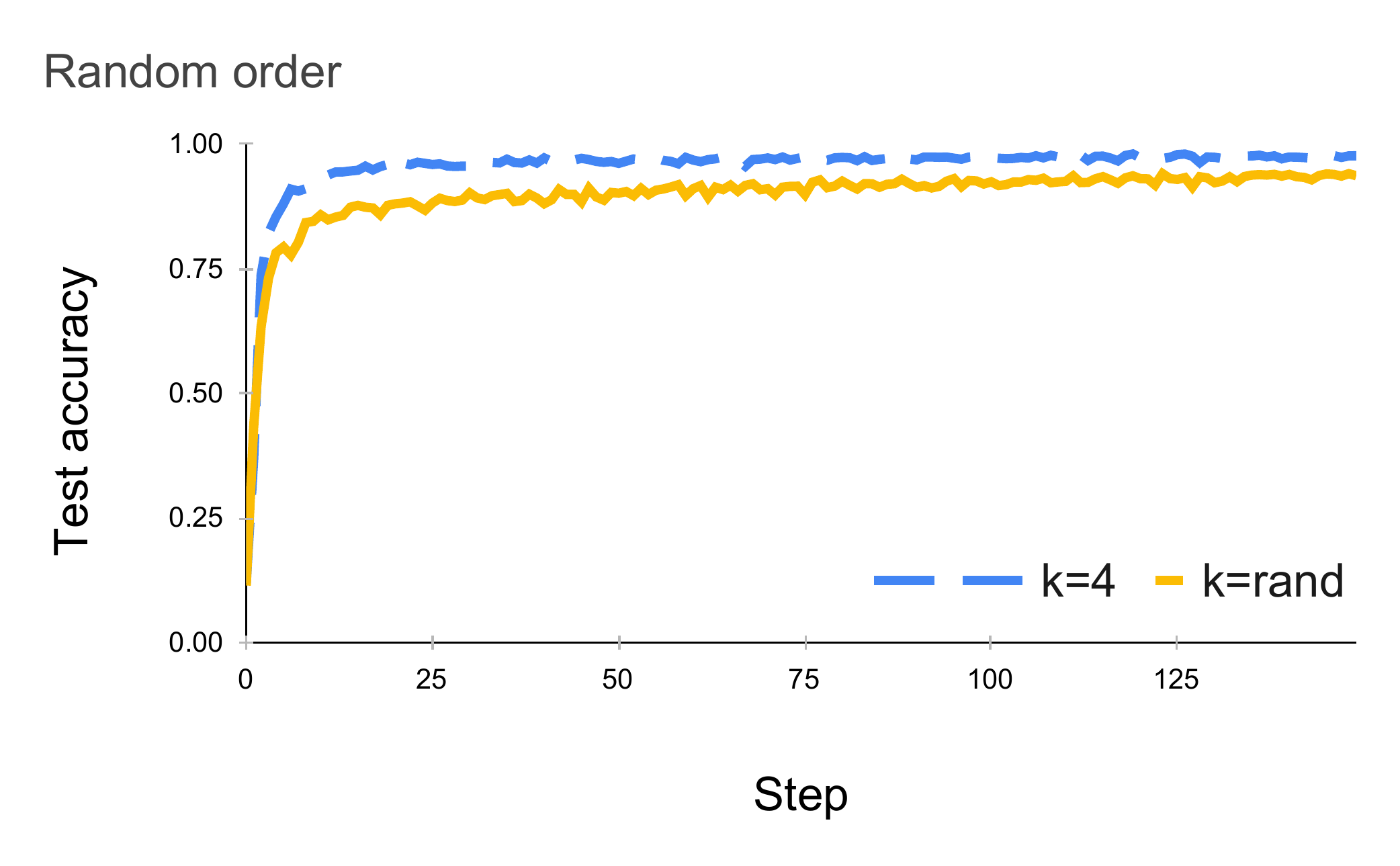}
    \caption{Learning curves for random number of variable update passes - for ``forward-backward'' and random order of operation updates.}
    \label{fig:attn_num_updates_comparison}
\end{figure}

\paragraph{Number of Energy Update Iterations}

In addition to comparing the number of energy update iterations $k$, we also try setting a random number of updates during training. 
Instead of specifying a fixed number of iterations $k$, we take a random $k$ between 1 and 5 and train the model with it. We evaluate the performance on inference with $k=3$ (the average value).
In \cref{fig:attn_num_updates_comparison} we compare the performance of the best model trained with random number of iterations $k$ with the best performing models trained with fixed $k$. 
As the plots show, the model trained with a random number of iterations performs on par with the best models with fixed $k$, but takes more time to train.


\section{Analysis of Alternative Initialization Strategies}\label{appendix:alternative_initializations}
In addition to the zero initialization for the output variable $y$, we also experiment with two more initialization strategies - random and uniform initialization. 
We perform this comparison for the MNIST experiment from \cref{section:experiments_mnist}.
For the random initialization, we initialize $y$ with random numbers from a uniform distribution on the interval $[0,1)$
and apply $\softmax$. For the uniform initialization, we assign equal values summing to one. 
We compare to the zero initialization strategy on the MNIST forward-backward experiment with $\gamma=.1$.
The results are presented in \cref{tab:alternative_initialization}. 
Random initialization shows promise, but the differences are small, and zero-init has the advantage of clearer parallels to the feed-forward case, so we report that and use it throughout all other experiments. The alternative initialization strategies can be further explored in further work.

\begin{table}[ht]
    \centering
    \caption{Comparison of different initialization strategies for the MNIST experiment from \cref{section:experiments_mnist}.}
    \vskip 0.15in
    \begin{sc}
    \begin{tabular}{cccccc}
        \toprule
         Initialization & $k=1$ & $k=2$ & $k=3$ & $k=4$ & $k=5$ \\
         \midrule
         zero & 98.75 & 98.74 & 98.83 & 98.78 & 98.69 \\
         random & 98.77 & 98.83 & 98.90 & 98.85 & 98.70 \\
         uniform & 98.76 & 98.74 & 98.83 & 98.78 & 98.68 \\
        \bottomrule
    \end{tabular}
    \end{sc}
    \label{tab:alternative_initialization}
\end{table}

\section{Results for forward-only UNN for MNIST}
In addition to the results for forward-backward training of the UNN, we also report results from training the UNN only in forward mode with $\gamma=0$, i.e. when the model is trained for image classification only. The results are in \cref{table:mnist_results_forward_only}. 

\begin{table}[h]
    \caption{MNIST accuracy with convolutional UNN in forward-only mode (\ie $\gamma=0$).}
    \vskip 0.15in
\begin{center}
\begin{small}
\begin{sc}
    \begin{tabular}{ll}
    \toprule
        Iterations & Accuracy \\
        \midrule
        $k=1,\gamma=0$ (baseline) & 98.80 \\
        $k=2$ & \textbf{98.82} \\
        $k=3$ & 98.75 \\
        $k=4$ & 98.74 \\
        $k=5$ & 98.69 \\
    \bottomrule
    \end{tabular}
\end{sc}
\end{small}
\end{center}
    \label{table:mnist_results_forward_only}
\end{table}

\section{Additional visualizations}\label{appendix:additional_viz}



\begin{figure*}[ht!]
    \centering
    \includegraphics[width=0.3\textwidth]{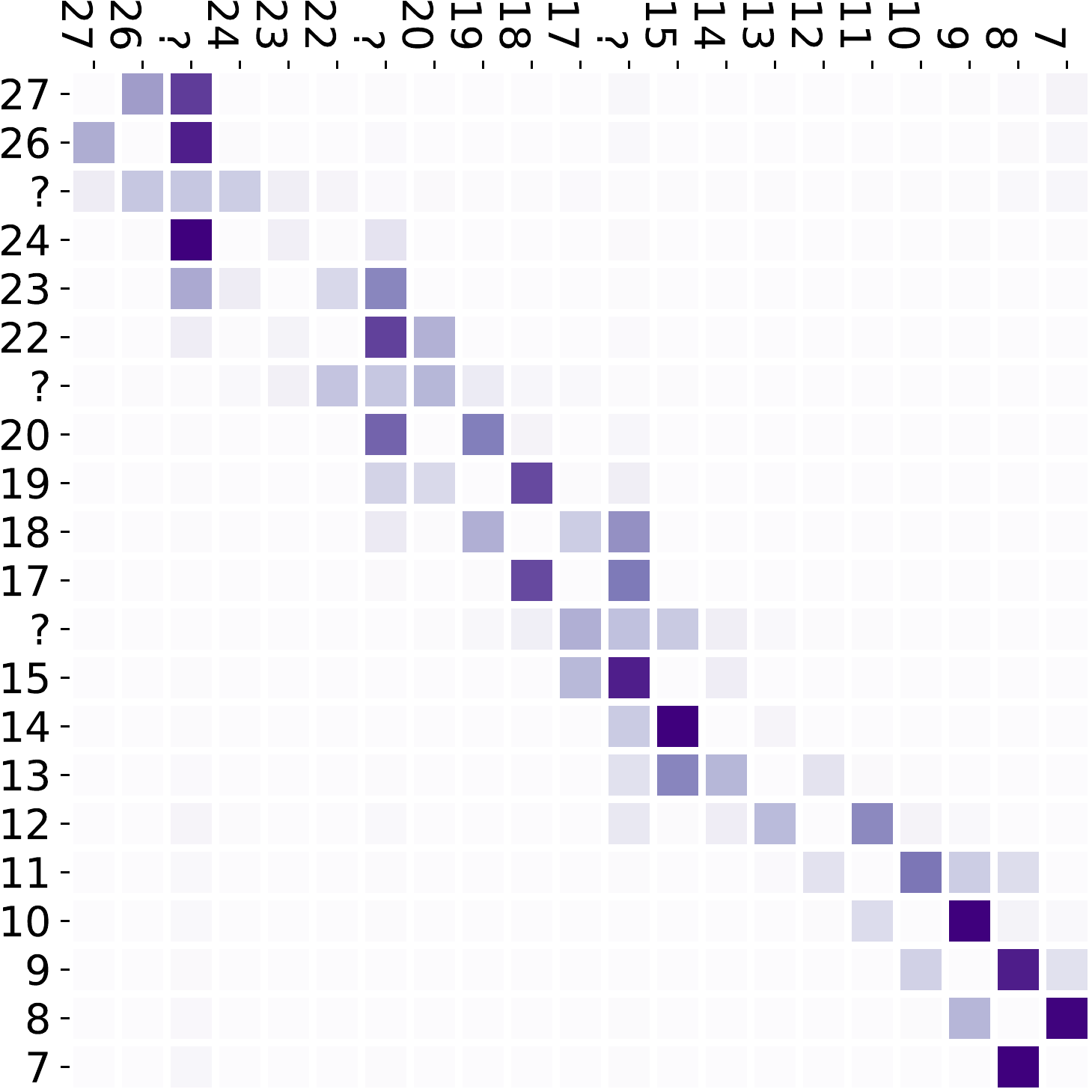}\hfill
    \includegraphics[width=0.3\textwidth]{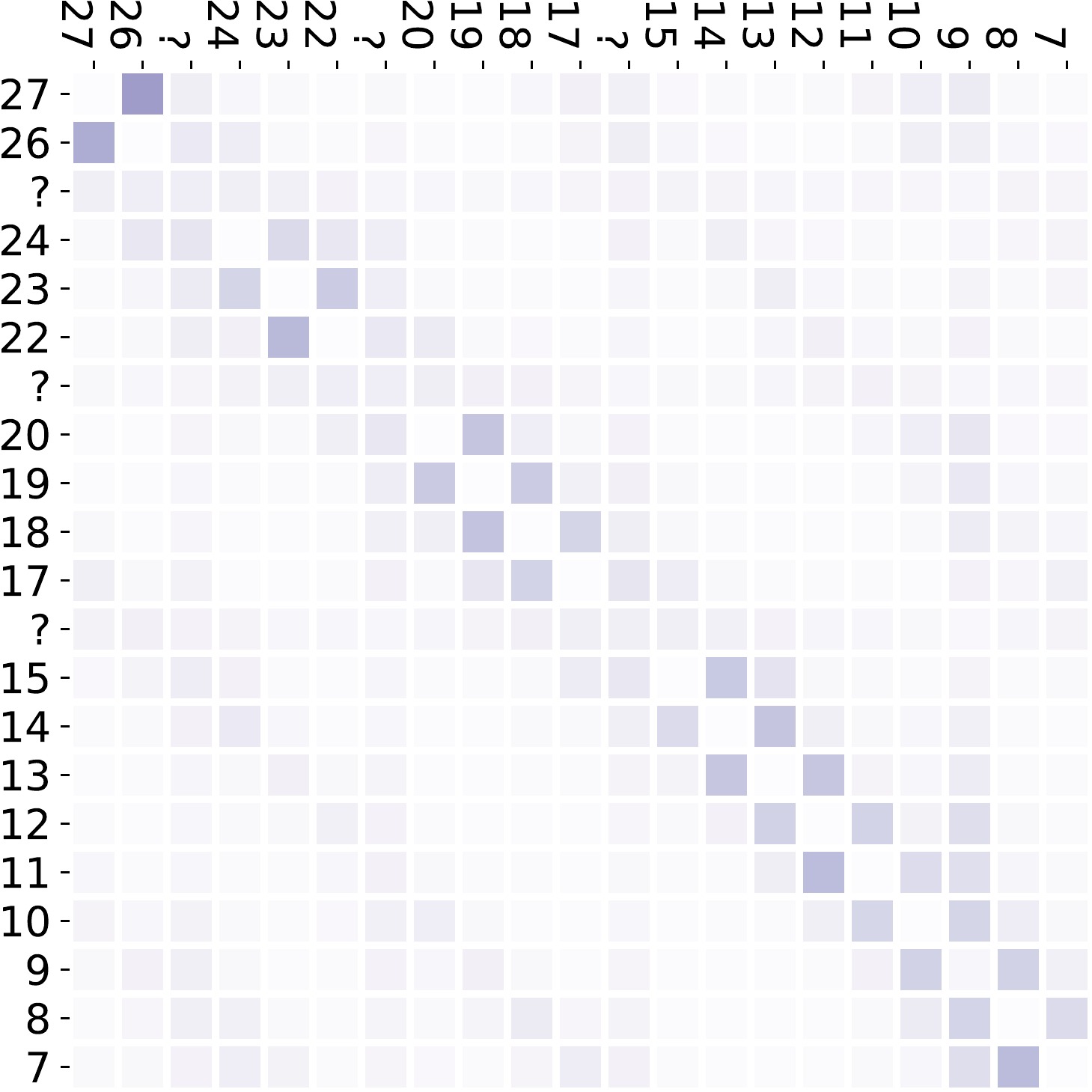}\hfill
    \includegraphics[width=0.3\textwidth]{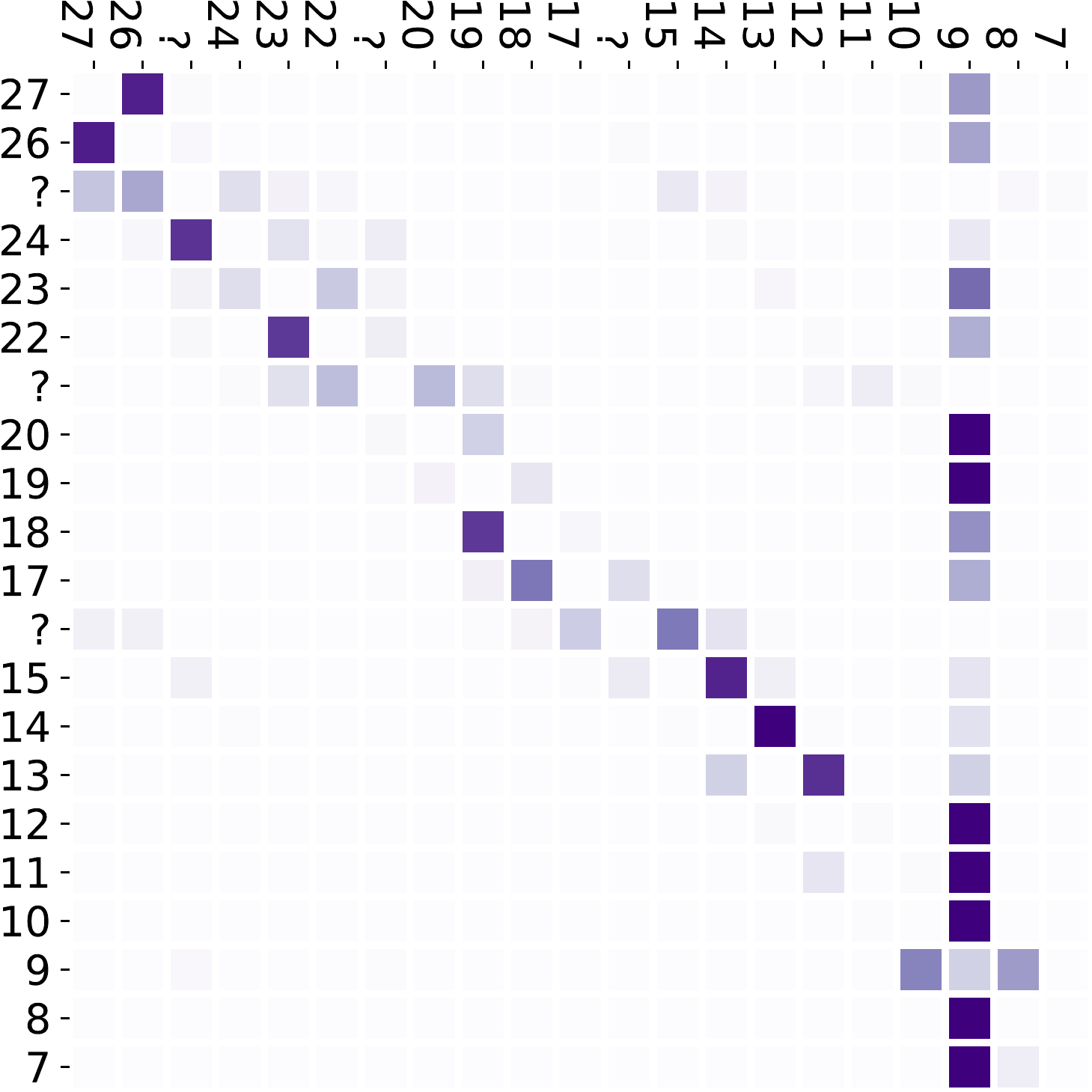}
    \\\medskip
    \includegraphics[width=0.32\textwidth]{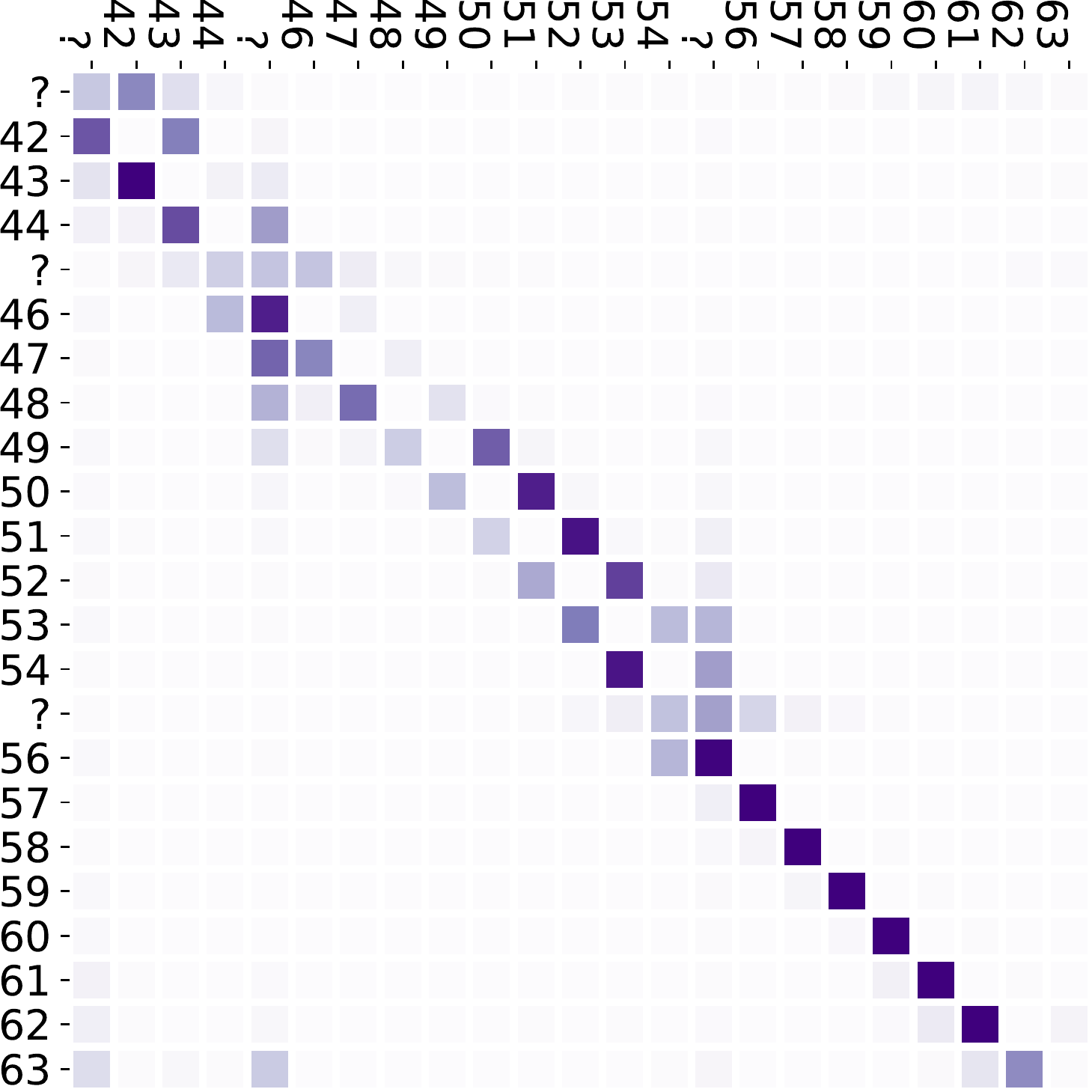}\hfill
    \includegraphics[width=0.32\textwidth]{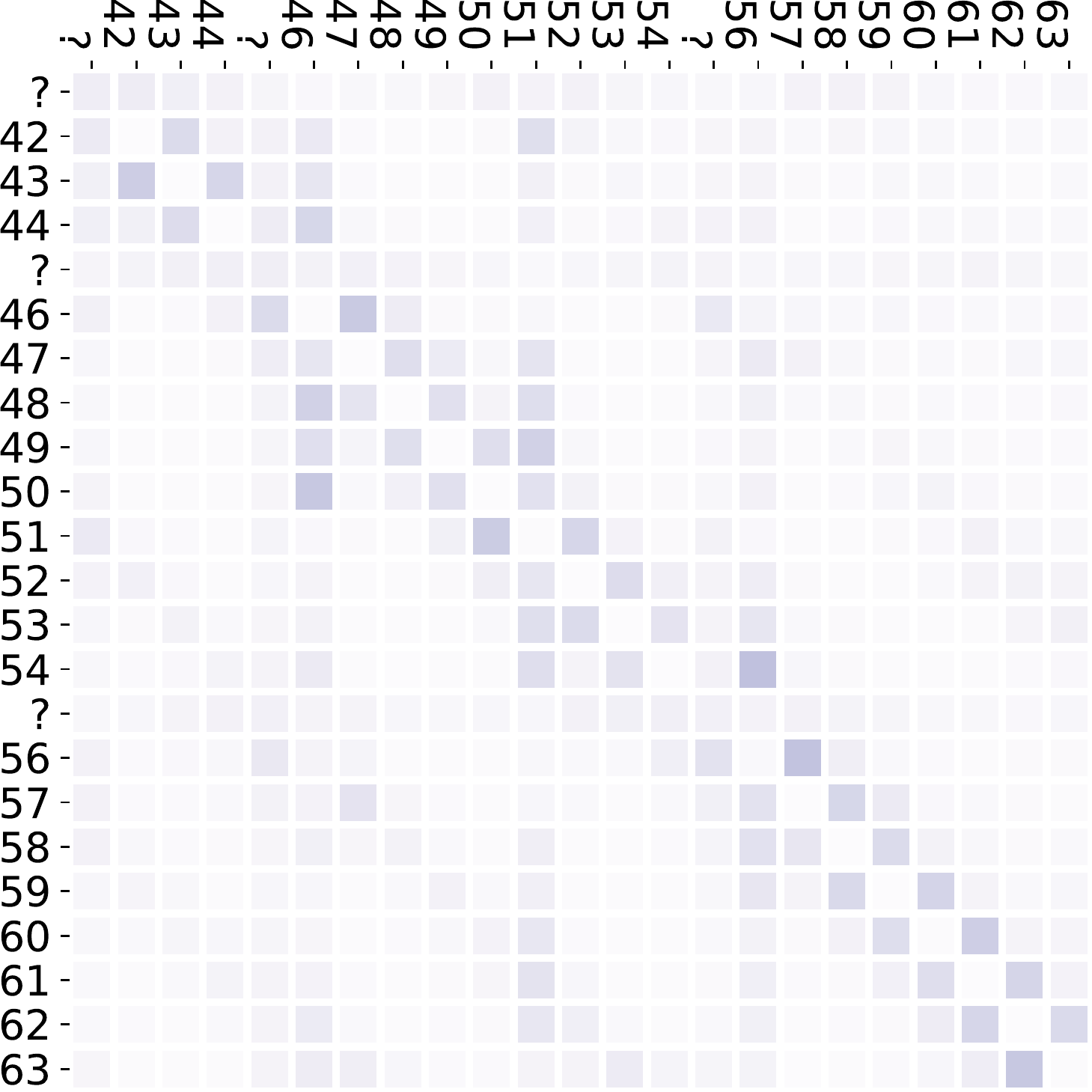}\hfill
    \includegraphics[width=0.32\textwidth]{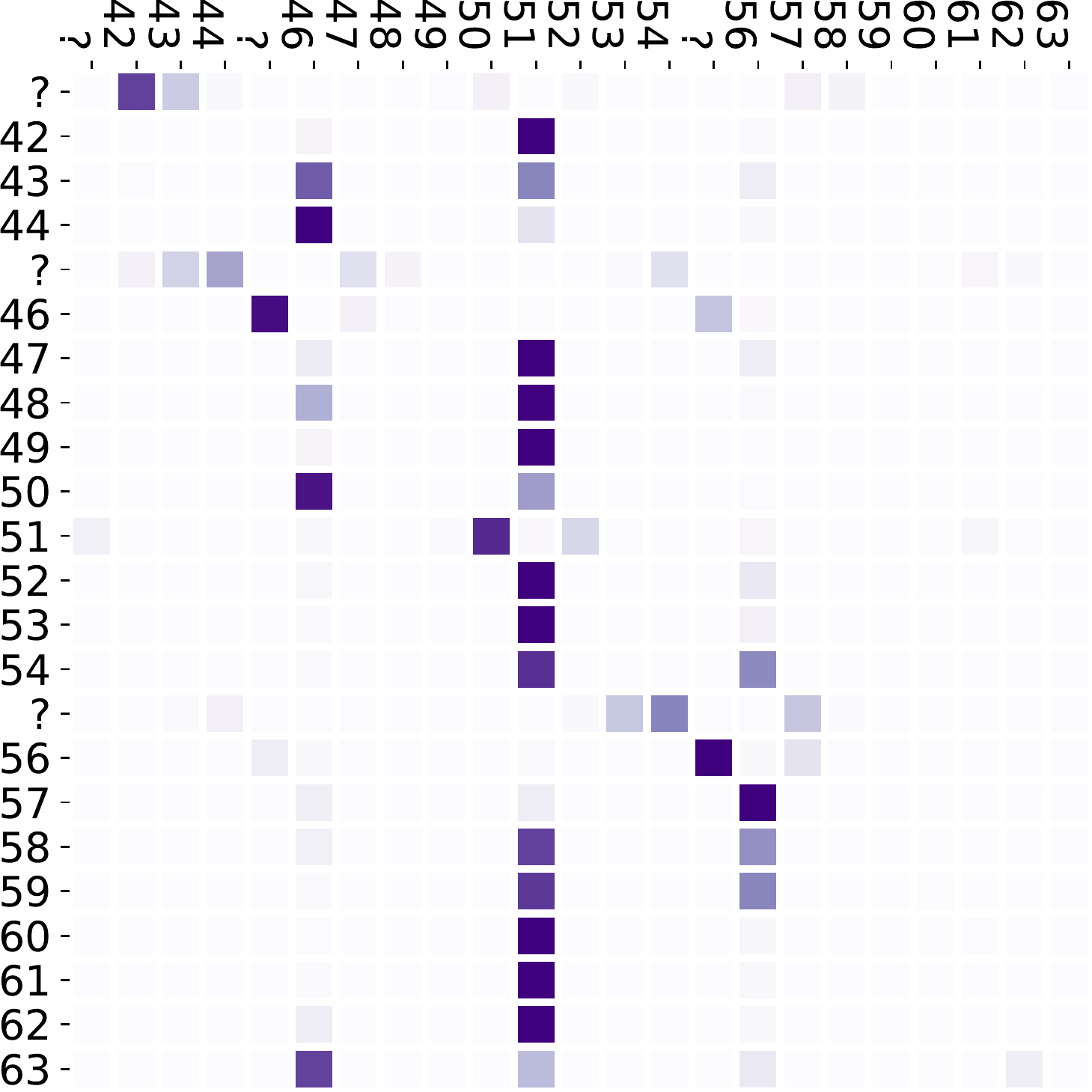}
    \\\medskip
    \includegraphics[width=0.29\textwidth]{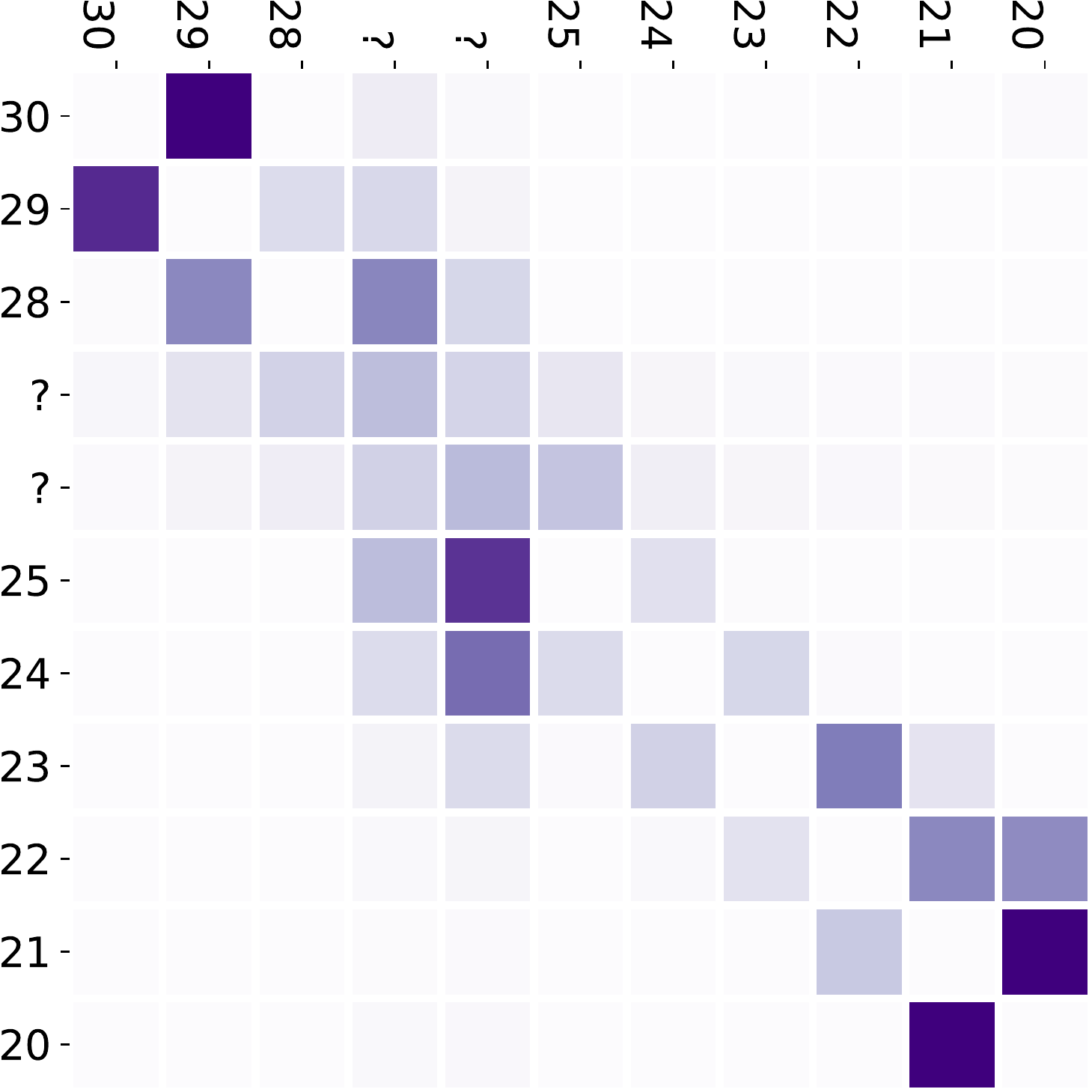}\hfill
    \includegraphics[width=0.29\textwidth]{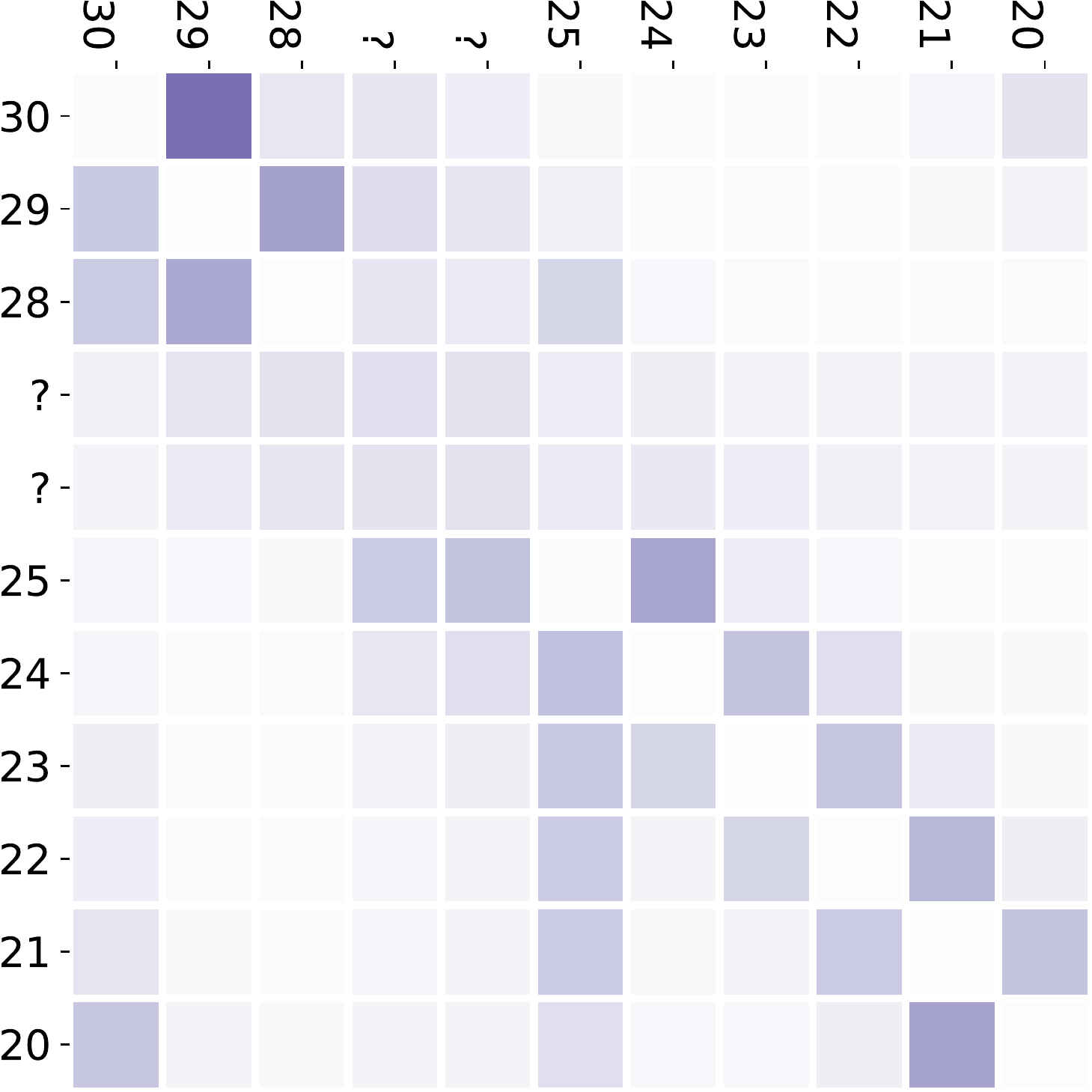}\hfill
    \includegraphics[width=0.34\textwidth]{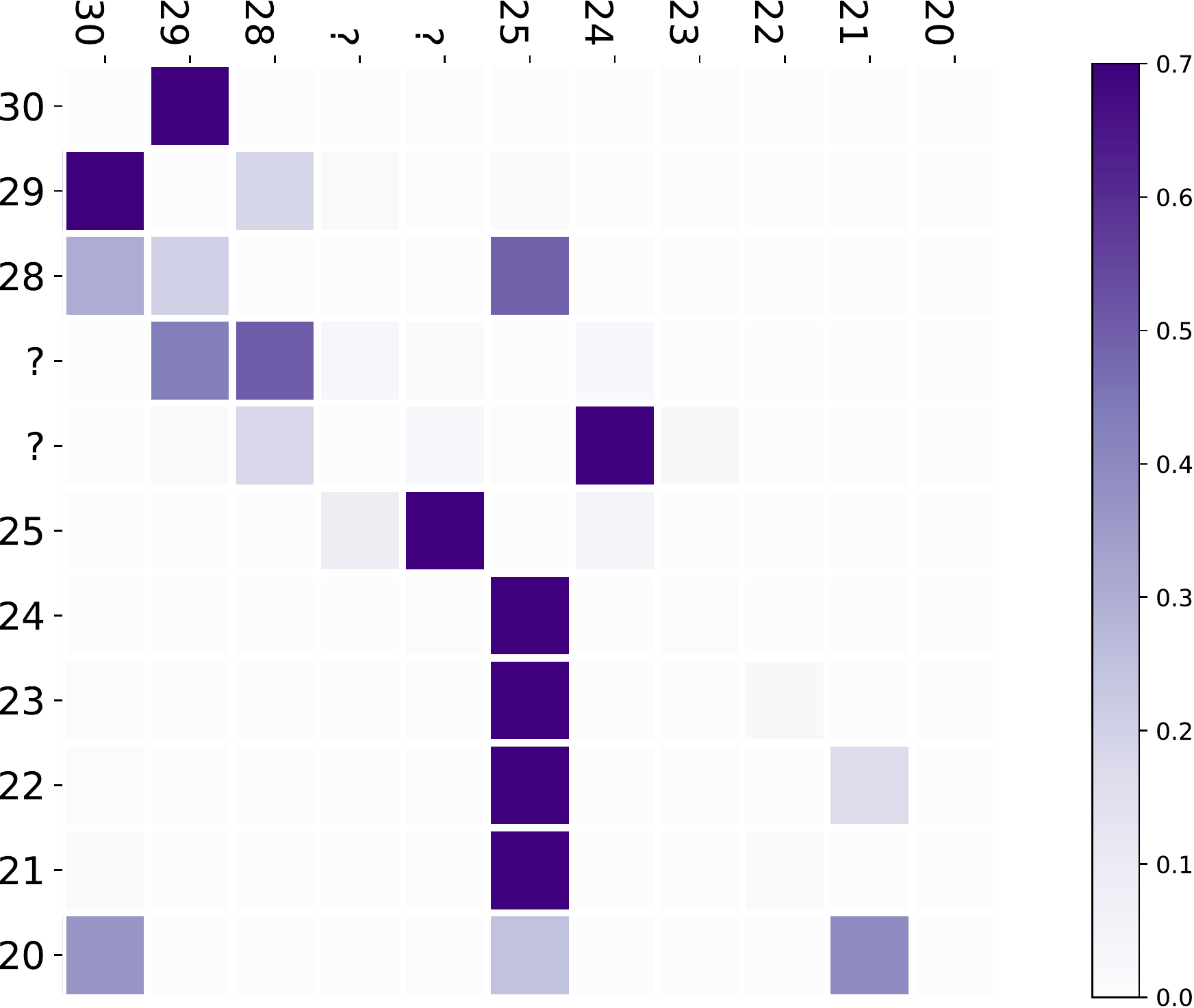}
    \caption{Example of the self-attention weights for models trained with $k=1$ (left) and $k=2$ after one iteration (middle) and two iterations (right). 
    For $k=2$, the model is more like an unrolled two-layer attention mechanism, with the first step identifying an off-diagonal pattern and the latter pooling information into an arbitrary token.
    }
    \label{fig:example_attn_weights}
\end{figure*}

\paragraph{Undirected self-attention weights.}
In \cref{fig:example_attn_weights} we show an example of the weights of the undirected self-attention described in \cref{section:experiments_attention}. The attention weights are the values of the variable $\var{S}$ calculated in the forward and backward pass. 

\begin{figure}
    \centering%
\def\H{.5}
\def\W{.9}
\begin{tikzpicture}[
    factoredge/.style={ultra thick,red},
    factor/.style={fill, rectangle, minimum width=2pt, minimum height=2pt},
    bv/.style={font=\small\tt}
    ]
\def\wa{undirected}
\def\wb{neural}
\def\wc{net}
\def\fdr{0}
\def\fdo{1}
\def\fdw{0}
\def\for{0}
\def\fod{0}
\def\fow{1}
\def\fwd{0}
\def\fwo{0}
\def\fwr{1}
\foreach \i in {1,...,3}{
\foreach \j in {1,...,3}{
    \coordinate (var \i-\j) at (\i*\W, \j*\H) {};
}
}
\pgfmathparse{int(40*\fdw)}
\node[bv, fill=red!\pgfmathresult] (dw) at (var 1-1) {\fdw};
\pgfmathparse{int(40*\fdo)}
\node[bv, fill=red!\pgfmathresult] (do) at (var 1-2) {\fdo};
\pgfmathparse{int(40*\fdr)}
\node[bv, fill=red!\pgfmathresult] (dr) at (var 1-3) {\fdr};
\pgfmathparse{int(40*\fow)}
\node[bv, fill=red!\pgfmathresult] (ow) at (var 2-1) {\fow};
\pgfmathparse{int(40*\fod)}
\node[bv, fill=red!\pgfmathresult] (od) at (var 2-3) {\fod};
\pgfmathparse{int(40*\for)}
\node[bv, fill=red!\pgfmathresult] (or) at (var 2-2) {\for};
\pgfmathparse{int(40*\fwo)}
\node[bv, fill=red!\pgfmathresult] (wo) at (var 3-2) {\fwo};
\pgfmathparse{int(40*\fwd)}
\node[bv, fill=red!\pgfmathresult] (wd) at (var 3-3) {\fwd};
\pgfmathparse{int(40*\fwr)}
\node[bv, fill=red!\pgfmathresult] (wr) at (var 3-1) {\fwr};
\node[left=10pt of var 1-3] {\small\texttt{\wa}$\to$};
\node[left=10pt of var 1-2] {\small\texttt{\wb}$\to$};
\node[left=10pt of var 1-1] {\small\texttt{\wc}$\to$};
\node[above left=20pt of var 1-3,rotate=-35,anchor=east]{\small\texttt{*}$\to$};
\node[above=10pt of var 1-3,anchor=south west,rotate=42] {\small\strut\texttt{\wa}};
\node[above=10pt of var 2-3,anchor=south west,rotate=42] {\small\strut\texttt{\wb}};
\node[above=10pt of var 3-3,anchor=south west,rotate=42] {\small\strut\texttt{\wc}};
\draw[rounded corners=3pt,thick,draw]
    ([xshift=-3pt,yshift=-3pt]dw.south west) rectangle
    ([xshift=3pt,yshift=3pt]wd.north east);
\end{tikzpicture}%
\begin{tikzpicture}[
    factoredge/.style={ultra thick,red},
    factor/.style={fill, rectangle, minimum width=2pt, minimum height=2pt},
    bv/.style={font=\small\tt}
    ]
\def\wa{undirected}
\def\wb{neural}
\def\wc{net}

\def\fdr{.02}
\def\fdo{.55}
\def\fdw{.43}
\def\for{.08}
\def\fod{.07}
\def\fow{.85}
\def\fwr{.90}
\def\fwd{.07}
\def\fwo{.03}
\coordinate (var 1 ne);
\foreach \i in {1,...,3}{
\foreach \j in {1,...,3}{
    \coordinate (var \i-\j) at (\i*\W, \j*\H) {};
}
}
\pgfmathparse{int(40*\fdw)}
\node[bv, fill=red!\pgfmathresult] (dw) at (var 1-1) {\fdw};
\pgfmathparse{int(40*\fdo)}
\node[bv, fill=red!\pgfmathresult] (do) at (var 1-2) {\fdo};
\pgfmathparse{int(40*\fdr)}
\node[bv, fill=red!\pgfmathresult] (dr) at (var 1-3) {\fdr};
\pgfmathparse{int(40*\fow)}
\node[bv, fill=red!\pgfmathresult] (ow) at (var 2-1) {\fow};
\pgfmathparse{int(40*\fod)}
\node[bv, fill=red!\pgfmathresult] (od) at (var 2-3) {\fod};
\pgfmathparse{int(40*\for)}
\node[bv, fill=red!\pgfmathresult] (or) at (var 2-2) {\for};
\pgfmathparse{int(40*\fwo)}
\node[bv, fill=red!\pgfmathresult] (wo) at (var 3-2) {\fwo};
\pgfmathparse{int(40*\fwd)}
\node[bv, fill=red!\pgfmathresult] (wd) at (var 3-3) {\fwd};
\pgfmathparse{int(40*\fwr)}
\node[bv, fill=red!\pgfmathresult] (wr) at (var 3-1) {\fwr};
\node[above=10pt of var 1-3,anchor=south west,rotate=42] {\small\strut\texttt{\wa}};
\node[above=10pt of var 2-3,anchor=south west,rotate=42] {\small\strut\texttt{\wb}};
\node[above=10pt of var 3-3,anchor=south west,rotate=42] {\small\strut\texttt{\wc}};
\draw[rounded corners=3pt,thick,draw]
    ([xshift=-3pt,yshift=-3pt]dw.south west) rectangle
    ([xshift=3pt,yshift=3pt]wd.north east);
\end{tikzpicture}
\caption{``Packed'' matrix representation of a dependency tree (left) and dependency arc marginals (right).
Each element corresponds to an arc \texttt{h$\rightarrow$m},
and the diagonal corresponds to the arcs from the root,
\texttt{*$\rightarrow$m}. The marginals, computed via the matrix-tree theorem,
are the structured counterpart of softmax, and correpond to arc probabilities.
}
\label{fig:example_marginals}
\end{figure}
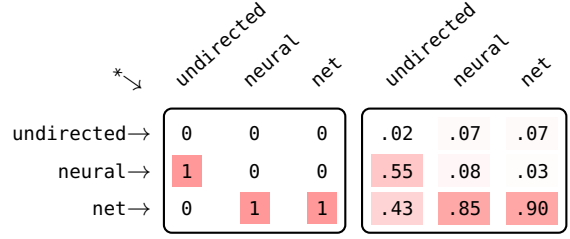

\paragraph{Dependency tree packed representation}
\Cref{fig:example_marginals} shows an example of how the trees are represented in the output of the model in \cref{section:experiments_parsing}. 

\section{Derivation of updates for convolutional UNN}
\label{section:conv_derivation}
The two-layer convolutional UNN is defined by the pairwise energies
\begin{equation}
\begin{aligned}
E_\var{XH_1}(X, H_1) &= -\DP{H_1}{\mathcal{C}_1(X; W_1)} \,, \\
E_\var{H_1H_2}(H_1, H_2) &= -\DP{H_2}{\mathcal{C}_2(H_1; W_2)}\,,\\
E_\var{H_2Y}(H_2, y) = -\DP{y}{VH_2}\,,
\end{aligned}
\end{equation}
and the unary energies
\begin{equation}
\begin{aligned}
E_\var{X}(X) &= \frac{1}{2}\|X\|_2^2 \,, \\
E_\var{H_1}(H_1) &= -\DP{H_1}{b_1 \otimes 1_{d_1}} + \Psi_{\tanh}(H_1)\,, \\
E_\var{H_2}(H_2) &= -\DP{H_2}{b_2 \otimes 1_{d_2}} + \Psi_{\tanh}(H_2)\,, \\
E_\var{Y}(y) &= -\DP{b}{y} - \entropy(y) \,. \\
\end{aligned}
\end{equation}
Above, $\mathcal{C}_1$ and $\mathcal{C}_2$ are 
are linear cross-correlation (convolution) operators with
stride two and filter weights $W_1 \in \reals^{32 \times 1 \times 6 \times 6}$ and $W_2 \in \reals^{64 \times 32 \times 4 \times 4}$, and
$b_1 \in \reals^{32}$ and $b_2 \in \reals^{64}$ are vectors of
biases for each convolutional filter. The hidden activations have dimension
$H_1 \in \reals^{32 \times (d_1)}$
and
$H_2 \in \reals^{64 \times (d_2)}$,
where $d_1$ and $d_2$ are tuples that depend on the input image size; for MNIST, $X \in \reals^{1 \times 28 \times 28}$ leading to $d_1=12 \times 12$ and $d_2=5 \times 5$.

To derive the energy updates, we use the fact that a real linear operator $\mathcal{A}$ interacts with the Frobenius inner product as:
\begin{equation}
\DP{P}{\mathcal{A}(Q)} = \DP{Q}{\mathcal{A}^\top(P)}\,,
\end{equation}
where $\mathcal{A}^\top$ is the transpose, or adjoint, operator.\footnote{This generalizes the observation that $p^\top A q = q^\top A^\top p$.}
If $\mathcal{C}$ is a convolution (\ie, \verb|torch.conv2d|)
then $\mathcal{C}^\top$ is a deconvolution (\ie, \verb|torch.conv_transpose2d|) with the same filters.
We then have
\begin{equation}
\begin{aligned}
E_\var{XH_1}(X, H_1) &= -\DP{H_1}{\mathcal{C}_1(X; W_1)} 
=-\DP{X}{\mathcal{C}_1^\top(H_1; W_1)} \,, \\
E_\var{H_1H_2}(H_1, H_2) &= -\DP{H_2}{\mathcal{C}_2(H_1; W_2)}
=-\DP{H_1}{\mathcal{C}_2^\top(H_2; W_2)}\,.\\
\end{aligned}
\end{equation}
Adding up all energies and ignoring the constant terms in each update, we get
\begin{equation}
\begin{aligned}
X_\star &= \argmin_X -\DP{X}{\mathcal{C}_1^\top(H_1, W_1)} + \Psi_{\tanh}(X) \\
&= \tanh(\mathcal{C}_1^\top(H_1, W_1))\,, \\
(H_1)_\star &= \argmin_{H_1} 
-\DP{H_1}{\mathcal{C}_1(X, W_1)} 
-\DP{H_1}{\mathcal{C}_2^\top(H_2, W_2)} 
-\DP{H_1}{b_1 \otimes 1_{d_1 \times d_1}}
+ \Psi_{\tanh(H_1)} \\
&= \tanh(
\mathcal{C}_1(X, W_1)+
\mathcal{C}_2^\top(H_2, W_2)+
b_1 \otimes 1_{d_1 \times d_1}
)\,, \\
(H_2)_\star &= \argmin_{H_2} 
-\DP{H_2}{\mathcal{C}_2(H_1, W_2)} 
-\DP{H_2}{\sigma_y(V)y}
-\DP{H_2}{b_2 \otimes 1_{d_2 \times d_2}}
+ \Psi_{\tanh(H_2)} \\
&= \tanh(
\mathcal{C}_2(H_1, W_2)
+\sigma_y(V)y
+b_2 \otimes 1_{d_2 \times d_2}
)\,, \\
y_\star &=
-\DP{y}{V H_2} - \DP{y}{b} - \entropy{y} \\
&= \softmax(VH_2 + b)\,.
\end{aligned}
\end{equation}
Note that in our case,
$H_2 \in \reals^{64 \times 5 \times 5}$,
$V \in \reals^{10 \times 64 \times 5 \times 5}$ and
thus $VH_2 \in \reals^{10}$ is a tensor contraction (\eg, \verb|torch.tensordot(V, H_2, dims=3)|).
The $\sigma_y$ linear operator -- opposite of $\rho$ from \cref{lemma:unrolling} -- rolls the axis of $V$ corresponding to $y$ to the \emph{last} position, such that
$\sigma_y(V) \in \reals^{64 \times 5 \times 5 \times 10}$, the tensor analogue of a transposition (\eg, \verb|torch.permute(V, (1, 2, 3, 0))|.)

%% file: main.bbl
\begin{thebibliography}{40}
\providecommand{\natexlab}[1]{#1}
\providecommand{\url}[1]{\texttt{#1}}
\expandafter\ifx\csname urlstyle\endcsname\relax
  \providecommand{\doi}[1]{doi: #1}\else
  \providecommand{\doi}{doi: \begingroup \urlstyle{rm}\Url}\fi

\bibitem[Ackley et~al.(1985)Ackley, Hinton, and Sejnowski]{ackley1985learning}
Ackley, D.~H., Hinton, G.~E., and Sejnowski, T.~J.
\newblock A learning algorithm for boltzmann machines.
\newblock \emph{Cognitive science}, 9\penalty0 (1):\penalty0 147--169, 1985.

\bibitem[Bahdanau et~al.(2014)Bahdanau, Cho, and Bengio]{bahdanau2014neural}
Bahdanau, D., Cho, K., and Bengio, Y.
\newblock Neural machine translation by jointly learning to align and
  translate.
\newblock \emph{arXiv preprint arXiv:1409.0473}, 2014.

\bibitem[Bak{\i}r et~al.(2007)Bak{\i}r, Hofmann, Smola, Sch{\"o}lkopf, and
  Taskar]{bakir2007predicting}
Bak{\i}r, G., Hofmann, T., Smola, A.~J., Sch{\"o}lkopf, B., and Taskar, B.
\newblock \emph{Predicting structured data}.
\newblock MIT press, 2007.

\bibitem[Belanger \& McCallum(2016)Belanger and
  McCallum]{belanger2016structured}
Belanger, D. and McCallum, A.
\newblock Structured prediction energy networks.
\newblock In \emph{International Conference on Machine Learning}, pp.\
  983--992. PMLR, 2016.

\bibitem[Belanger et~al.(2017)Belanger, Yang, and McCallum]{belanger2017end}
Belanger, D., Yang, B., and McCallum, A.
\newblock End-to-end learning for structured prediction energy networks.
\newblock In \emph{International Conference on Machine Learning}, pp.\
  429--439. PMLR, 2017.

\bibitem[Chu \& Liu(1965)Chu and Liu]{Chu1965}
Chu, Y.-J. and Liu, T.-H.
\newblock On the shortest arborescence of a directed graph.
\newblock \emph{Science Sinica}, 14:\penalty0 1396--1400, 1965.

\bibitem[Dehghani et~al.(2018)Dehghani, Gouws, Vinyals, Uszkoreit, and
  Kaiser]{dehghani2018universal}
Dehghani, M., Gouws, S., Vinyals, O., Uszkoreit, J., and Kaiser, L.
\newblock Universal transformers.
\newblock In \emph{International Conference on Learning Representations}, 2018.

\bibitem[Deng(2012)]{deng2012mnist}
Deng, L.
\newblock The mnist database of handwritten digit images for machine learning
  research.
\newblock \emph{IEEE Signal Processing Magazine}, 29\penalty0 (6):\penalty0
  141--142, 2012.

\bibitem[Domke(2012)]{domke2012optimizationbasedlearning}
Domke, J.
\newblock Generic methods for optimization-based modeling.
\newblock In \emph{Artificial Intelligence and Statistics}, pp.\  318--326.
  PMLR, 2012.

\bibitem[Dozat \& Manning(2016)Dozat and Manning]{dozat2016deep}
Dozat, T. and Manning, C.~D.
\newblock Deep biaffine attention for neural dependency parsing.
\newblock \emph{arXiv preprint arXiv:1611.01734}, 2016.

\bibitem[Du \& Mordatch(2019)Du and Mordatch]{du_mordatch}
Du, Y.-l. and Mordatch, I.
\newblock Implicit generation and modeling with energy based models.
\newblock In \emph{NeurIPS}, 2019.

\bibitem[Duvenaud et~al.(2020)Duvenaud, Kolter, and Johnson]{duvenaud2020deep}
Duvenaud, D., Kolter, J.~Z., and Johnson, M.
\newblock Deep implicit layers tutorial-neural odes, deep equilibirum models,
  and beyond.
\newblock \emph{Neural Information Processing Systems Tutorial}, 2020.

\bibitem[Edmonds(1967)]{Edmonds1967}
Edmonds, J.
\newblock \href{https://doi.org/10.6028%2Fjres.071b.032}{Optimum branchings}.
\newblock \emph{J. Res. Nat. Bur. Stand.}, 71B:\penalty0 233--240, 1967.

\bibitem[Grathwohl et~al.(2020{\natexlab{a}})Grathwohl, Wang, Jacobsen,
  Duvenaud, Norouzi, and Swersky]{Grathwohl2020Your}
Grathwohl, W., Wang, K.-C., Jacobsen, J.-H., Duvenaud, D., Norouzi, M., and
  Swersky, K.
\newblock Your classifier is secretly an energy based model and you should
  treat it like one.
\newblock In \emph{International Conference on Learning Representations},
  2020{\natexlab{a}}.

\bibitem[Grathwohl et~al.(2020{\natexlab{b}})Grathwohl, Wang, Jacobsen,
  Duvenaud, and Zemel]{grathwohl2020learning}
Grathwohl, W., Wang, K.-C., Jacobsen, J.-H., Duvenaud, D., and Zemel, R.
\newblock Learning the stein discrepancy for training and evaluating
  energy-based models without sampling.
\newblock In \emph{ICML}. PMLR, 2020{\natexlab{b}}.

\bibitem[Hershey et~al.(2014)Hershey, Roux, and
  Weninger]{hershey2014deepunfolding}
Hershey, J.~R., Roux, J.~L., and Weninger, F.
\newblock Deep unfolding: Model-based inspiration of novel deep architectures.
\newblock \emph{arXiv preprint arXiv:1409.2574}, 2014.

\bibitem[Hinton(2007)]{hinton2007boltzmann}
Hinton, G.~E.
\newblock Boltzmann machine.
\newblock \emph{Scholarpedia}, 2\penalty0 (5):\penalty0 1668, 2007.

\bibitem[Hopfield(1984)]{hopfield1984neurons}
Hopfield, J.~J.
\newblock Neurons with graded response have collective computational properties
  like those of two-state neurons.
\newblock \emph{Proceedings of the national academy of sciences}, 81\penalty0
  (10):\penalty0 3088--3092, 1984.

\bibitem[Kingma \& Ba(2015)Kingma and Ba]{kingma2014adam}
Kingma, D.~P. and Ba, J.
\newblock Adam: {A} method for stochastic optimization.
\newblock In \emph{3rd International Conference on Learning Representations,
  {ICLR} 2015, San Diego, CA, USA, May 7-9, 2015, Conference Track
  Proceedings}, 2015.
\newblock URL \url{http://arxiv.org/abs/1412.6980}.

\bibitem[Kiperwasser \& Goldberg(2016)Kiperwasser and
  Goldberg]{kiperwasser-goldberg-2016-simple}
Kiperwasser, E. and Goldberg, Y.
\newblock Simple and accurate dependency parsing using bidirectional {LSTM}
  feature representations.
\newblock \emph{Transactions of the Association for Computational Linguistics},
  4:\penalty0 313--327, 2016.
\newblock \doi{10.1162/tacl_a_00101}.
\newblock URL \url{https://www.aclweb.org/anthology/Q16-1023}.

\bibitem[Kirchhoff(1847)]{Kirchhoff1847}
Kirchhoff, G.
\newblock Ueber die aufl{\"o}sung der gleichungen, auf welche man bei der
  untersuchung der linearen vertheilung galvanischer str{\"o}me gef{\"u}hrt
  wird.
\newblock \emph{Annalen der Physik}, 148\penalty0 (12):\penalty0 497--508,
  1847.

\bibitem[Koo et~al.(2007)Koo, Globerson, Carreras~P{\'e}rez, and
  Collins]{koo2007structured}
Koo, T., Globerson, A., Carreras~P{\'e}rez, X., and Collins, M.
\newblock Structured prediction models via the matrix-tree theorem.
\newblock In \emph{Joint Conference on Empirical Methods in Natural Language
  Processing and Computational Natural Language Learning (EMNLP-CoNLL)}, pp.\
  141--150, 2007.

\bibitem[K{\"u}bler et~al.(2009)K{\"u}bler, McDonald, and
  Nivre]{kubler2009dependency}
K{\"u}bler, S., McDonald, R., and Nivre, J.
\newblock Dependency parsing.
\newblock \emph{Synthesis lectures on human language technologies}, 1\penalty0
  (1):\penalty0 1--127, 2009.

\bibitem[Lawson et~al.(2019)Lawson, Tucker, Dai, and
  Ranganath]{lawson2019energyinspiredmodels}
Lawson, J., Tucker, G., Dai, B., and Ranganath, R.
\newblock Energy-inspired models: Learning with sampler-induced distributions.
\newblock In Wallach, H., Larochelle, H., Beygelzimer, A., d\textquotesingle
  Alch\'{e}-Buc, F., Fox, E., and Garnett, R. (eds.), \emph{Advances in Neural
  Information Processing Systems}, volume~32. Curran Associates, Inc., 2019.
\newblock URL
  \url{https://proceedings.neurips.cc/paper/2019/file/28659414dab9eca0219dd592b8136434-Paper.pdf}.

\bibitem[LeCun et~al.(2006)LeCun, Chopra, Hadsell, Ranzato, and
  Huang]{lecun2006tutorial}
LeCun, Y., Chopra, S., Hadsell, R., Ranzato, M., and Huang, F.
\newblock A tutorial on energy-based learning.
\newblock \emph{Predicting structured data}, 1\penalty0 (0), 2006.

\bibitem[Li \& Zemel(2014)Li and Zemel]{li2014meanfieldnetworks}
Li, Y. and Zemel, R.~S.
\newblock Mean-field networks.
\newblock \emph{ArXiv}, abs/1410.5884, 2014.

\bibitem[Martins et~al.(2009)Martins, Smith, and Xing]{martins2009polyhedral}
Martins, A.~F., Smith, N.~A., and Xing, E.~P.
\newblock Polyhedral outer approximations with application to natural language
  parsing.
\newblock In \emph{Proceedings of the 26th Annual International Conference on
  Machine Learning}, pp.\  713--720, 2009.

\bibitem[McDonald \& Satta(2007)McDonald and Satta]{mcdonald2007complexity}
McDonald, R. and Satta, G.
\newblock On the complexity of non-projective data-driven dependency parsing.
\newblock In \emph{Proceedings of the Tenth International Conference on Parsing
  Technologies}, pp.\  121--132, 2007.

\bibitem[Ngiam et~al.(2011)Ngiam, Chen, Koh, and Ng]{ngiam2011learning}
Ngiam, J., Chen, Z., Koh, P.~W., and Ng, A.~Y.
\newblock Learning deep energy models.
\newblock In \emph{ICML}, 2011.

\bibitem[Nowozin et~al.(2014)Nowozin, Gehler, Jancsary, and
  Lampert]{nowozin2014advanced}
Nowozin, S., Gehler, P.~V., Jancsary, J., and Lampert, C.~H.
\newblock \emph{Advanced Structured Prediction}.
\newblock MIT Press, 2014.

\bibitem[Ramsauer et~al.(2020)Ramsauer, Sch{\"a}fl, Lehner, Seidl, Widrich,
  Gruber, Holzleitner, Pavlovi{\'c}, Sandve, Greiff,
  et~al.]{ramsauer2020hopfield}
Ramsauer, H., Sch{\"a}fl, B., Lehner, J., Seidl, P., Widrich, M., Gruber, L.,
  Holzleitner, M., Pavlovi{\'c}, M., Sandve, G.~K., Greiff, V., et~al.
\newblock Hopfield networks is all you need.
\newblock \emph{arXiv preprint arXiv:2008.02217}, 2020.

\bibitem[Smith \& Smith(2007)Smith and Smith]{smith2007probabilistic}
Smith, D.~A. and Smith, N.~A.
\newblock Probabilistic models of nonprojective dependency trees.
\newblock In \emph{Proceedings of the 2007 Joint Conference on Empirical
  Methods in Natural Language Processing and Computational Natural Language
  Learning (EMNLP-CoNLL)}, pp.\  132--140, 2007.

\bibitem[Smith(2011)]{smith2011linguistic}
Smith, N.~A.
\newblock Linguistic structure prediction.
\newblock \emph{Synthesis lectures on human language technologies}, 4\penalty0
  (2):\penalty0 1--274, 2011.

\bibitem[Smolensky(1986)]{smolensky1986information}
Smolensky, P.
\newblock Information processing in dynamical systems: Foundations of harmony
  theory.
\newblock Technical report, Colorado Univ at Boulder Dept of Computer Science,
  1986.

\bibitem[Tu et~al.(2020)Tu, Pang, and Gimpel]{tu-etal-2020-improving}
Tu, L., Pang, R.~Y., and Gimpel, K.
\newblock Improving joint training of inference networks and structured
  prediction energy networks.
\newblock In \emph{Proceedings of the Fourth Workshop on Structured Prediction
  for NLP}, pp.\  62--73, Online, November 2020. Association for Computational
  Linguistics.
\newblock \doi{10.18653/v1/2020.spnlp-1.8}.
\newblock URL \url{https://aclanthology.org/2020.spnlp-1.8}.

\bibitem[Vaswani et~al.(2017)Vaswani, Shazeer, Parmar, Uszkoreit, Jones, Gomez,
  Kaiser, and Polosukhin]{vaswani2017attention}
Vaswani, A., Shazeer, N., Parmar, N., Uszkoreit, J., Jones, L., Gomez, A.~N.,
  Kaiser, {\L}., and Polosukhin, I.
\newblock Attention is all you need.
\newblock In \emph{Advances in Neural Information Processing Systems}, 2017.

\bibitem[Wainwright \& Jordan(2008)Wainwright and
  Jordan]{wainwright2008graphical}
Wainwright, M.~J. and Jordan, M.~I.
\newblock \emph{Graphical models, exponential families, and variational
  inference}.
\newblock Now Publishers Inc, 2008.

\bibitem[Welling et~al.(2004)Welling, Rosen-Zvi, and
  Hinton]{welling2004exponential}
Welling, M., Rosen-Zvi, M., and Hinton, G.
\newblock Exponential family harmoniums with an application to information
  retrieval.
\newblock In \emph{Proceedings of the 17th International Conference on Neural
  Information Processing Systems}, 2004.

\bibitem[Xu \& Yin(2013)Xu and Yin]{xu2013block}
Xu, Y. and Yin, W.
\newblock A block coordinate descent method for regularized multiconvex
  optimization with applications to nonnegative tensor factorization and
  completion.
\newblock \emph{SIAM Journal on Imaging Sciences}, 6\penalty0 (3):\penalty0
  1758--1789, 2013.

\bibitem[Zeman et~al.(2020)Zeman, Nivre, Abrams, Ackermann, Aepli, Aghaei,
  Agi{\'c}, Ahmadi, Ahrenberg, Ajede, Aleksandravi{\v c}i{\=u}t{\.e}, Alfina,
  Antonsen, Aplonova, Aquino, Aragon, Aranzabe, Arnard{\'o}ttir, Arutie,
  Arwidarasti, Asahara, Ateyah, Atmaca, Attia, Atutxa, Augustinus, Badmaeva,
  Balasubramani, Ballesteros, Banerjee, Bank, Barbu~Mititelu, Basmov,
  Batchelor, Bauer, Bedir, Bengoetxea, Berk, Berzak, Bhat, Bhat, Biagetti,
  Bick, Bielinskien{\.e}, Bjarnad{\'o}ttir, Blokland, Bobicev, Boizou,
  Borges~V{\"o}lker, B{\"o}rstell, Bosco, Bouma, Bowman, Boyd, Brokait{\.e},
  Burchardt, Candito, Caron, Caron, Cavalcanti, Cebiro{\u g}lu~Eryi{\u g}it,
  Cecchini, Celano, {\v C}{\'e}pl{\"o}, Cetin, {\c C}etino{\u g}lu, Chalub,
  Chi, Cho, Choi, Chun, Cignarella, Cinkov{\'a}, Collomb, {\c C}{\"o}ltekin,
  Connor, Courtin, Davidson, de~Marneffe, de~Paiva, Derin, de~Souza, Diaz~de
  Ilarraza, Dickerson, Dinakaramani, Dione, Dirix, Dobrovoljc, Dozat,
  Droganova, Dwivedi, Eckhoff, Eli, Elkahky, Ephrem, Erina, Erjavec, Etienne,
  Evelyn, Facundes, Farkas, Fernanda, Fernandez~Alcalde, Foster, Freitas,
  Fujita, Gajdo{\v s}ov{\'a}, Galbraith, Garcia, G{\"a}rdenfors, Garza,
  Gerardi, Gerdes, Ginter, Goenaga, Gojenola, G{\"o}k{\i}rmak, Goldberg,
  G{\'o}mez~Guinovart, Gonz{\'a}lez~Saavedra, Grici{\=u}t{\.e}, Grioni, Grobol,
  Gr{\= u}z{\={\i}}tis, Guillaume, Guillot-Barbance, G{\"u}ng{\"o}r, Habash,
  Hafsteinsson, Haji{\v c}, Haji{\v c}~jr., H{\"a}m{\"a}l{\"a}inen,
  H{\`a}~M{\~y}, Han, Hanifmuti, Hardwick, Harris, Haug, Heinecke, Hellwig,
  Hennig, Hladk{\'a}, Hlav{\'a}{\v c}ov{\'a}, Hociung, Hohle, Huber, Hwang,
  Ikeda, Ingason, Ion, Irimia, Ishola, Jel{\'{\i}}nek, Johannsen,
  J{\'o}nsd{\'o}ttir, J{\o}rgensen, Juutinen, K, Ka{\c s}{\i}kara, Kaasen,
  Kabaeva, Kahane, Kanayama, Kanerva, Katz, Kayadelen, Kenney, Kettnerov{\'a},
  Kirchner, Klementieva, K{\"o}hn, K{\"o}ksal, Kopacewicz, Korkiakangas,
  Kotsyba, Kovalevskait{\.e}, Krek, Krishnamurthy, Kwak, Laippala, Lam,
  Lambertino, Lando, Larasati, Lavrentiev, Lee, L{\^e}~H{\`{\^o}}ng, Lenci,
  Lertpradit, Leung, Levina, Li, Li, Li, Li, Lim, Lind{\'e}n, Ljube{\v
  s}i{\'c}, Loginova, Luthfi, Luukko, Lyashevskaya, Lynn, Macketanz,
  Makazhanov, Mandl, Manning, Manurung, M{\u a}r{\u a}nduc, Mare{\v c}ek,
  Marheinecke, Mart{\'{\i}}nez~Alonso, Martins, Ma{\v s}ek, Matsuda, Matsumoto,
  {McDonald}, {McGuinness}, Mendon{\c c}a, Miekka, Mischenkova, Misirpashayeva,
  Missil{\"a}, Mititelu, Mitrofan, Miyao, Mojiri~Foroushani, Moloodi,
  Montemagni, More, Moreno~Romero, Mori, Mori, Morioka, Moro, Mortensen,
  Moskalevskyi, Muischnek, Munro, Murawaki, M{\"u}{\"u}risep, Nainwani,
  Nakhl{\'e}, Navarro~Hor{\~n}iacek, Nedoluzhko, Ne{\v s}pore-B{\=e}rzkalne,
  Nguy{\~{\^e}}n~Th{\d i}, Nguy{\~{\^e}}n Th{\d i}~Minh, Nikaido, Nikolaev,
  Nitisaroj, Nourian, Nurmi, Ojala, Ojha, Ol{\'u}{\`o}kun, Omura, Onwuegbuzia,
  Osenova, {\"O}stling, {\O}vrelid, {\"O}zate{\c s}, {\"O}zg{\"u}r,
  {\"O}zt{\"u}rk~Ba{\c s}aran, Partanen, Pascual, Passarotti, Patejuk,
  Paulino-Passos, Peljak-{\L}api{\'n}ska, Peng, Perez, Perkova, Perrier,
  Petrov, Petrova, Phelan, Piitulainen, Pirinen, Pitler, Plank, Poibeau,
  Ponomareva, Popel, Pretkalni{\c n}a, Pr{\'e}vost, Prokopidis,
  Przepi{\'o}rkowski, Puolakainen, Pyysalo, Qi, R{\"a}{\"a}bis, Rademaker,
  Rama, Ramasamy, Ramisch, Rashel, Rasooli, Ravishankar, Real, Rebeja, Reddy,
  Rehm, Riabov, Rie{\ss}ler, Rimkut{\.e}, Rinaldi, Rituma, Rocha,
  R{\"o}gnvaldsson, Romanenko, Rosa, Roșca, Rovati, Rudina, Rueter,
  R{\'u}narsson, Sadde, Safari, Sagot, Sahala, Saleh, Salomoni, Samard{\v
  z}i{\'c}, Samson, Sanguinetti, S{\"a}rg, Saul{\={\i}}te, Sawanakunanon,
  Scannell, Scarlata, Schneider, Schuster, Seddah, Seeker, Seraji, Shen,
  Shimada, Shirasu, Shohibussirri, Sichinava, Sigurdsson, Silveira, Silveira,
  Simi, Simionescu, Simk{\'o}, {\v S}imkov{\'a}, Simov, Skachedubova, Smith,
  Soares-Bastos, Spadine, Steingr{\'{\i}}msson, Stella, Straka, Strickland,
  Strnadov{\'a}, Suhr, Sulestio, Sulubacak, Suzuki, Sz{\'a}nt{\'o}, Taji,
  Takahashi, Tamburini, Tan, Tanaka, Tella, Tellier, Thomas, Torga, Toska,
  Trosterud, Trukhina, Tsarfaty, T{\"u}rk, Tyers, Uematsu, Untilov, Ure{\v
  s}ov{\'a}, Uria, Uszkoreit, Utka, Vajjala, van Niekerk, van Noord, Varga,
  Villemonte de~la Clergerie, Vincze, Wakasa, Wallenberg, Wallin, Walsh, Wang,
  Washington, Wendt, Widmer, Williams, Wir{\'e}n, Wittern, Woldemariam, Wong,
  Wr{\'o}blewska, Yako, Yamashita, Yamazaki, Yan, Yasuoka, Yavrumyan, Yu, {\v
  Z}abokrtsk{\'y}, Zahra, Zeldes, Zhu, and
  Zhuravleva]{universal-dependencies-2.7}
Zeman, D., Nivre, J., Abrams, M., Ackermann, E., Aepli, N., Aghaei, H.,
  Agi{\'c}, {\v Z}., Ahmadi, A., Ahrenberg, L., Ajede, C.~K., Aleksandravi{\v
  c}i{\=u}t{\.e}, G., Alfina, I., Antonsen, L., Aplonova, K., Aquino, A.,
  Aragon, C., Aranzabe, M.~J., Arnard{\'o}ttir, {\t H}., Arutie, G.,
  Arwidarasti, J.~N., Asahara, M., Ateyah, L., Atmaca, F., Attia, M., Atutxa,
  A., Augustinus, L., Badmaeva, E., Balasubramani, K., Ballesteros, M.,
  Banerjee, E., Bank, S., Barbu~Mititelu, V., Basmov, V., Batchelor, C., Bauer,
  J., Bedir, S.~T., Bengoetxea, K., Berk, G., Berzak, Y., Bhat, I.~A., Bhat,
  R.~A., Biagetti, E., Bick, E., Bielinskien{\.e}, A., Bjarnad{\'o}ttir, K.,
  Blokland, R., Bobicev, V., Boizou, L., Borges~V{\"o}lker, E., B{\"o}rstell,
  C., Bosco, C., Bouma, G., Bowman, S., Boyd, A., Brokait{\.e}, K., Burchardt,
  A., Candito, M., Caron, B., Caron, G., Cavalcanti, T., Cebiro{\u g}lu~Eryi{\u
  g}it, G., Cecchini, F.~M., Celano, G. G.~A., {\v C}{\'e}pl{\"o}, S., Cetin,
  S., {\c C}etino{\u g}lu, {\"O}., Chalub, F., Chi, E., Cho, Y., Choi, J.,
  Chun, J., Cignarella, A.~T., Cinkov{\'a}, S., Collomb, A., {\c C}{\"o}ltekin,
  {\c C}., Connor, M., Courtin, M., Davidson, E., de~Marneffe, M.-C., de~Paiva,
  V., Derin, M.~O., de~Souza, E., Diaz~de Ilarraza, A., Dickerson, C.,
  Dinakaramani, A., Dione, B., Dirix, P., Dobrovoljc, K., Dozat, T., Droganova,
  K., Dwivedi, P., Eckhoff, H., Eli, M., Elkahky, A., Ephrem, B., Erina, O.,
  Erjavec, T., Etienne, A., Evelyn, W., Facundes, S., Farkas, R., Fernanda, M.,
  Fernandez~Alcalde, H., Foster, J., Freitas, C., Fujita, K., Gajdo{\v
  s}ov{\'a}, K., Galbraith, D., Garcia, M., G{\"a}rdenfors, M., Garza, S.,
  Gerardi, F.~F., Gerdes, K., Ginter, F., Goenaga, I., Gojenola, K.,
  G{\"o}k{\i}rmak, M., Goldberg, Y., G{\'o}mez~Guinovart, X.,
  Gonz{\'a}lez~Saavedra, B., Grici{\=u}t{\.e}, B., Grioni, M., Grobol, L.,
  Gr{\= u}z{\={\i}}tis, N., Guillaume, B., Guillot-Barbance, C.,
  G{\"u}ng{\"o}r, T., Habash, N., Hafsteinsson, H., Haji{\v c}, J., Haji{\v
  c}~jr., J., H{\"a}m{\"a}l{\"a}inen, M., H{\`a}~M{\~y}, L., Han, N.-R.,
  Hanifmuti, M.~Y., Hardwick, S., Harris, K., Haug, D., Heinecke, J., Hellwig,
  O., Hennig, F., Hladk{\'a}, B., Hlav{\'a}{\v c}ov{\'a}, J., Hociung, F.,
  Hohle, P., Huber, E., Hwang, J., Ikeda, T., Ingason, A.~K., Ion, R., Irimia,
  E., Ishola, {\d O}., Jel{\'{\i}}nek, T., Johannsen, A., J{\'o}nsd{\'o}ttir,
  H., J{\o}rgensen, F., Juutinen, M., K, S., Ka{\c s}{\i}kara, H., Kaasen, A.,
  Kabaeva, N., Kahane, S., Kanayama, H., Kanerva, J., Katz, B., Kayadelen, T.,
  Kenney, J., Kettnerov{\'a}, V., Kirchner, J., Klementieva, E., K{\"o}hn, A.,
  K{\"o}ksal, A., Kopacewicz, K., Korkiakangas, T., Kotsyba, N.,
  Kovalevskait{\.e}, J., Krek, S., Krishnamurthy, P., Kwak, S., Laippala, V.,
  Lam, L., Lambertino, L., Lando, T., Larasati, S.~D., Lavrentiev, A., Lee, J.,
  L{\^e}~H{\`{\^o}}ng, P., Lenci, A., Lertpradit, S., Leung, H., Levina, M.,
  Li, C.~Y., Li, J., Li, K., Li, Y., Lim, K., Lind{\'e}n, K., Ljube{\v
  s}i{\'c}, N., Loginova, O., Luthfi, A., Luukko, M., Lyashevskaya, O., Lynn,
  T., Macketanz, V., Makazhanov, A., Mandl, M., Manning, C., Manurung, R., M{\u
  a}r{\u a}nduc, C., Mare{\v c}ek, D., Marheinecke, K., Mart{\'{\i}}nez~Alonso,
  H., Martins, A., Ma{\v s}ek, J., Matsuda, H., Matsumoto, Y., {McDonald}, R.,
  {McGuinness}, S., Mendon{\c c}a, G., Miekka, N., Mischenkova, K.,
  Misirpashayeva, M., Missil{\"a}, A., Mititelu, C., Mitrofan, M., Miyao, Y.,
  Mojiri~Foroushani, A., Moloodi, A., Montemagni, S., More, A., Moreno~Romero,
  L., Mori, K.~S., Mori, S., Morioka, T., Moro, S., Mortensen, B.,
  Moskalevskyi, B., Muischnek, K., Munro, R., Murawaki, Y., M{\"u}{\"u}risep,
  K., Nainwani, P., Nakhl{\'e}, M., Navarro~Hor{\~n}iacek, J.~I., Nedoluzhko,
  A., Ne{\v s}pore-B{\=e}rzkalne, G., Nguy{\~{\^e}}n~Th{\d i}, L.,
  Nguy{\~{\^e}}n Th{\d i}~Minh, H., Nikaido, Y., Nikolaev, V., Nitisaroj, R.,
  Nourian, A., Nurmi, H., Ojala, S., Ojha, A.~K., Ol{\'u}{\`o}kun, A., Omura,
  M., Onwuegbuzia, E., Osenova, P., {\"O}stling, R., {\O}vrelid, L.,
  {\"O}zate{\c s}, {\c S}.~B., {\"O}zg{\"u}r, A., {\"O}zt{\"u}rk~Ba{\c s}aran,
  B., Partanen, N., Pascual, E., Passarotti, M., Patejuk, A., Paulino-Passos,
  G., Peljak-{\L}api{\'n}ska, A., Peng, S., Perez, C.-A., Perkova, N., Perrier,
  G., Petrov, S., Petrova, D., Phelan, J., Piitulainen, J., Pirinen, T.~A.,
  Pitler, E., Plank, B., Poibeau, T., Ponomareva, L., Popel, M., Pretkalni{\c
  n}a, L., Pr{\'e}vost, S., Prokopidis, P., Przepi{\'o}rkowski, A.,
  Puolakainen, T., Pyysalo, S., Qi, P., R{\"a}{\"a}bis, A., Rademaker, A.,
  Rama, T., Ramasamy, L., Ramisch, C., Rashel, F., Rasooli, M.~S., Ravishankar,
  V., Real, L., Rebeja, P., Reddy, S., Rehm, G., Riabov, I., Rie{\ss}ler, M.,
  Rimkut{\.e}, E., Rinaldi, L., Rituma, L., Rocha, L., R{\"o}gnvaldsson, E.,
  Romanenko, M., Rosa, R., Roșca, V., Rovati, D., Rudina, O., Rueter, J.,
  R{\'u}narsson, K., Sadde, S., Safari, P., Sagot, B., Sahala, A., Saleh, S.,
  Salomoni, A., Samard{\v z}i{\'c}, T., Samson, S., Sanguinetti, M., S{\"a}rg,
  D., Saul{\={\i}}te, B., Sawanakunanon, Y., Scannell, K., Scarlata, S.,
  Schneider, N., Schuster, S., Seddah, D., Seeker, W., Seraji, M., Shen, M.,
  Shimada, A., Shirasu, H., Shohibussirri, M., Sichinava, D., Sigurdsson,
  E.~F., Silveira, A., Silveira, N., Simi, M., Simionescu, R., Simk{\'o}, K.,
  {\v S}imkov{\'a}, M., Simov, K., Skachedubova, M., Smith, A., Soares-Bastos,
  I., Spadine, C., Steingr{\'{\i}}msson, S., Stella, A., Straka, M.,
  Strickland, E., Strnadov{\'a}, J., Suhr, A., Sulestio, Y.~L., Sulubacak, U.,
  Suzuki, S., Sz{\'a}nt{\'o}, Z., Taji, D., Takahashi, Y., Tamburini, F., Tan,
  M. A.~C., Tanaka, T., Tella, S., Tellier, I., Thomas, G., Torga, L., Toska,
  M., Trosterud, T., Trukhina, A., Tsarfaty, R., T{\"u}rk, U., Tyers, F.,
  Uematsu, S., Untilov, R., Ure{\v s}ov{\'a}, Z., Uria, L., Uszkoreit, H.,
  Utka, A., Vajjala, S., van Niekerk, D., van Noord, G., Varga, V., Villemonte
  de~la Clergerie, E., Vincze, V., Wakasa, A., Wallenberg, J.~C., Wallin, L.,
  Walsh, A., Wang, J.~X., Washington, J.~N., Wendt, M., Widmer, P., Williams,
  S., Wir{\'e}n, M., Wittern, C., Woldemariam, T., Wong, T.-s., Wr{\'o}blewska,
  A., Yako, M., Yamashita, K., Yamazaki, N., Yan, C., Yasuoka, K., Yavrumyan,
  M.~M., Yu, Z., {\v Z}abokrtsk{\'y}, Z., Zahra, S., Zeldes, A., Zhu, H., and
  Zhuravleva, A.
\newblock Universal dependencies 2.7, 2020.
\newblock URL \url{http://hdl.handle.net/11234/1-3424}.
\newblock {LINDAT}/{CLARIAH}-{CZ} digital library at the Institute of Formal
  and Applied Linguistics ({{\'U}FAL}), Faculty of Mathematics and Physics,
  Charles University.

\end{thebibliography}
